%% file: main.tex
\pdfoutput=1
\documentclass{article}

\usepackage{microtype}
\usepackage{graphicx}
\usepackage{subfigure}
\usepackage{hyperref}

\usepackage{xcolor}
\usepackage{siunitx}

\usepackage{float} % for placing the huge tables in the appendix at the right place

\usepackage{tabularray}
\UseTblrLibrary{booktabs}

\usepackage{natbib}

\usepackage[accepted]{icml2024}

\usepackage{amsmath}
\usepackage{amssymb}
\usepackage{mathtools}
\usepackage{amsthm}

\usepackage[capitalize,noabbrev]{cleveref}

\theoremstyle{plain}
\newtheorem{theorem}{Theorem}[section]
\newtheorem{proposition}[theorem]{Proposition}

\newtheorem{corollary}[theorem]{Corollary}
\theoremstyle{definition}

\theoremstyle{remark}

\usepackage[textsize=tiny]{todonotes}

\icmltitlerunning{Adversarial Circuit Evaluation}

\newcommand{\dataset}{\mathcal{D}}
\newcommand{\datasetcorrupted}{\widetilde{\dataset}}

\DeclareMathOperator{\Unif}{Unif} % uniform distribution
\DeclareMathOperator{\kldivop}{D_{KL}}
\newcommand{\kldiv}[2]{\kldivop\left(#1 \parallel #2 \right)} % this suggests a better solution that scales with larger symbols: https://tex.stackexchange.com/questions/151984/double-vertical-bar-notation

\newcommand{\tempval}[1]{\texttt{[#1]}} % this suggests a better solution that scales with larger symbols: https://tex.stackexchange.com/questions/151984/double-vertical-bar-notation

\newcommand{\newlinesymbol}{\textbackslash n}
\newcommand{\bossymbol}{}

\DeclareMathOperator{\Prob}{\mathcal{P}}
\DeclareMathOperator{\Binom}{Binom}
\DeclareMathOperator{\Bernoulli}{Bern}
\DeclarePairedDelimiter{\ceil}{\lceil}{\rceil}

\begin{document}

\twocolumn[
\icmltitle{Adversarial Circuit Evaluation}



\icmlsetsymbol{equal}{*}

\begin{icmlauthorlist}
\icmlauthor{Niels uit de Bos}{indep}
\icmlauthor{Adri\`a Garriga-Alonso}{far}
\end{icmlauthorlist}

\icmlaffiliation{indep}{Independent. Work done at MATS}
\icmlaffiliation{far}{FAR AI, San Diego, CA, United States}

\icmlcorrespondingauthor{Niels uit de Bos}{niels@uit-de-bos.nl}

\icmlkeywords{Mechanistic Interpretability, Circuits, Evaluation}

\vskip 0.3in
]



\printAffiliationsAndNotice{}  % leave blank if no need to mention equal contribution

\begin{abstract}
   Circuits are supposed to accurately describe how a neural network performs a specific task,
   but do they really? 
   We evaluate three circuits found in the literature (IOI, greater-than, and docstring)
   in an adversarial manner, considering inputs where the
   circuit's behavior maximally diverges from the full model.
   Concretely, we measure the KL divergence between the full model's output and the circuit's output,
   calculated through resample ablation,
   and we analyze the worst-performing inputs.
   Our results show that
   the circuits for the IOI and docstring tasks fail to behave similarly to the full model
   even on completely benign inputs from the original task,
   indicating that more robust circuits are needed for safety-critical applications.
\end{abstract}

\section{Introduction}

Neural networks' vast size and complexity make them difficult to reverse engineer.
To address this issue, circuits have been proposed \cite{olah2020zoom} as one paradigm.
By isolating the subset of components of a neural network that perform a chosen, narrow task,
we hope to obtain a subnetwork
that is smaller and disentangled from all other tasks the full network performs,
making it easier to understand.
We call this subset of components the \emph{circuit}.

For the circuit to be helpful in understanding the original, full model,
it is crucial that the circuit's behavior coincides with the full model's behavior
on the chosen task.
In particular, on task-specific inputs,
the circuit should produce the same output distribution as the full model.
Previous work has mostly assessed a circuit's performance by
testing its ability to output the same distribution as the full model
\emph{on average} on the task-specific inputs.

In this paper, we argue that, besides looking at the average performance,
it is worthwhile to assess a circuit by analyzing its \emph{worst-case} performance:
on which inputs and how large a proportion of inputs
does the circuit fail to emulate the full model's behavior?
We propose a method to evaluate circuits from this adversarial perspective
and apply it to several circuits found in the literature:
Indirect Object Identification (IOI) \cite{wangInterpretabilityWildCircuit2022},
greater-than \cite{hannaHowDoesGPT22023},
and docstring \cite{heimersheimCircuitPythonDocstrings2023}.

Adversarial circuit evaluation is important for several reasons.
First, we cannot say we to truly understand the full model if the circuit behaves
differently on a certain fraction of the inputs.
For example, our analysis (\cref{sec:results}) showsthat the circuit for the IOI task
fails to emulate the full model's behavior on a significant fraction of inputs. We speculate that especially when romantic objects are involved,
components outside of the circuit play a crucial role.
Since romantic objects could be only a small fraction of the inputs tested,
the circuit's average performance can be high,
even though we may be missing a crucial piece of the puzzle.

In particular, if we ever want to use circuits for guarantees or in safety-critical applications, it is crucial to
describe the neural network's behavior on all inputs.
As a general principle,
evolutionary pressures and adversarial attacks can successfully discover and exploit edge cases.
For example, if we edit a reward model to align more with human values, the policy optimizing for it may
find a regime where our edits fail.
In the absence of specific circuit-based safety interventions,
this remains speculation. However, the benign-seeming yet adversarial examples
we find in this paper might convince the reader that any safety measure could
only be built with more robust circuits.

Moreover, we argue that the adversarial metrics are not only useful for
evaluating circuits but also for improving them.
Our analysis of the worst-performing inputs for the circuits
(\crefrange{table:ioi-outputs}{table:greaterthan-outputs})
shows failure modes that a researcher could try to inspect and address manually.
Alternatively, our adversarial evaluation metrics
could be plugged in to automatic circuit discovery techniques,
likely leading to more robust circuits.

Our main contributions are the following:
\begin{itemize}
   \item We provide a method to calculate the proposed adversarial metrics (\cref{sec:methodology}).
      \footnote{Our code is available on GitHub at \url{https://github.com/Nielius/AdversarialCircuitEvaluation}.}
   \item We prove a formula to calculate how many task data points are needed
      to bound the circuit's worst-case performance with high probability (\cref{sec:how-many-samples-needed});
      in more technical terms, we calculate the sample size required to find high-probability upper bounds for
      percentiles (e.g., the 99th percentile) of the KL divergence between the circuit's output and the full model's output
      on a distribution of task data points.
   \item We identify subtasks of IOI and Docstring where the circuit especially fails to explain
      (\cref{sec:results}; in particular, \cref{table:ioi-outputs} and \cref{table:docstring-outputs}). 
      For IOI, the circuit fails most strongly on inputs featuring a romantic object (``kiss'' or ``necklace'');
      for Docstring, patch inputs that have \texttt{file} as one of their parameters
      disrupt the circuit's performance, causing it to predict \texttt{file} as the next parameter, even though it does
      not appear in the clean input.
   \item In contrast, we find the Greater-Than circuit is more robust than the other two
      and does not have exhibit any significant edge case failures (\cref{sec:results} and \cref{table:greaterthan-outputs}).
\end{itemize}

\section{Methodology}
\label{sec:methodology}

The core component of our adversarial circuit evaluation method
is the calculation of the KL divergence between the circuit's output
and the full model's output for a large sample of input points.
The adversarial metrics we are interested in extracting
are the maximum KL divergence and several high percentile values from
the resulting distribution of KL divergences.
In this section, we describe the technical details of this calculation.

\paragraph{Resample ablation}
Let $\dataset$ denote the set of inputs for the task.
To calculate the output for the circuit, we ablate all components
not in the circuit.
While any kind of ablation could be used,
we use \emph{resample ablation} in our experiments.
That means that in addition to the dataset $\dataset$ of inputs for the task,
we need a dataset $\datasetcorrupted$ of \emph{corrupted inputs}
(sometimes called \emph{patch inputs}). On a forward pass with ablations, we replace the output of an ablated component
with its output from the (unablated) forward pass on a corrupted input.
Resample ablation,
also sometimes known as \emph{patching with different activations} (e.g. \cite{conmyAutomatedCircuitDiscovery2023}),
is a kind of \emph{interchange intervention} \cite{geigerCausalAbstractionFaithful2023};
more background and justification can also be found in \cite{chan2022causal} and \cite{zhang2024best}.
The circuits analyzed in this paper were identified through resample ablation.

For the reader's benefit, we explain in more detail
how resample ablation works in the context of the circuits analyzed in this paper.
The neural networks we are dealing with are transformers,
and the circuits are subsets of edges between nodes that represent MLPs, transformer heads (each individually), the embedded input,
and the output before the unembedding.
The transformer heads have three different inputs: the key, the query, and the value.
Because of the residual stream's additivity,
each node can causally impact all downstream nodes,
so there is an edge from each node to all downstream nodes.
If an edge from a node $X$ to a node $Y$ is not part of the circuit $C$
and we want to calculate the output $C(x, \tilde x)$ of the circuit $C$ on a clean input $x$
with a corrupted input $\tilde x$,
then $X$'s contribution $X(x)$ to the input of $Y$ in a forward pass on $x$ is replaced with
$X$'s contribution $X(\tilde x)$ to the input of $Y$ from a forward pass on $\tilde x$.
In our implementation, we achieve this by subtracting $X(x)$ from the input to $Y$, and adding $X(\tilde x)$;
this works because of the residual stream's additivity.
In this way, intuitively speaking,
a component can output clean outputs to some of its dependent downstream components,
while simultaneously outputting corrupted outputs to other components.

\paragraph{Evaluation metrics}
For $x \in \dataset$, we denote by $M(x)$ the output of the full model $M$ on $x$. For $x \in \dataset, \tilde x \in \datasetcorrupted$,
we denote by $C(x, \tilde x)$ the output of a forward pass of $M$
where we have resample-ablated all components outside of the circuit $C$
using the corrupted input $\tilde x$.
The outputs $C(x, \tilde x)$ and $M(x)$ are categorical probability distributions
over the model vocabulary. We use the KL divergence
$\kldiv{M(x)}{C(x, \tilde x)}$ to measure how close the circuit's output is to the model's output.
To obtain our adversarial evaluation metrics, we sample a sufficiently
large number of points from the distribution
\begin{equation}
   \begin{split}
   \kldiv{M(x)}{C(x, \tilde x)}
   \text{ for } x &\sim \Unif(\dataset),  \\
   \tilde x &\sim \Unif(\datasetcorrupted).
   \end{split}
   \label{eq:kl-div}
\end{equation}
We then take the maximum or high percentiles of this distribution
of KL divergences
as our adversarial evaluation metric.

\subsection{How many samples are needed?}
\label{sec:how-many-samples-needed}

The method described above samples from a distribution
and then takes a percentile from that sample.
How close is the sample percentile to the true percentile,
and how many points do you need to sample to get a good estimate?

The following result, which we prove in \cref{sec:proof-percentile-bounds},
provides an answer.
Let $X$ be a real-valued probability distribution
and let $0 < p < 1$.
Denote by $x_p$ the true $p$-th percentile of $X$.
Because we are looking at worst-case scenarios, we would like a tight upper bound $\hat x_p$
for $x_p$.
Take any $\epsilon > 0$ with $p + \epsilon < 1$
and write $\hat x_p$ for the $\lceil (p + \epsilon) \cdot n \rceil$-th order
statistic of $n$ i.i.d samples from $X$, i.e.,
the $\lceil (p + \epsilon) \cdot n \rceil$-th smallest value of the $n$ i.i.d. samples.
We then have the following result.

\begin{proposition}
   \label{thm:percentile-bound-as-binomial}
   The probability $\Pr(\hat x_p \geq x_p)$ that $\hat x_p$ is an upper bound for the
   true $p$-th percentile $x_p$ of $X$ can be calculated as
   \begin{equation}
      \Pr\left(\hat x_p \geq x_p \right) = F_{\rm Binom}(\lceil (p + \epsilon) \cdot n \rceil - 1; n, p)
      \label{eq:percentile-binomial-calculation}
   \end{equation}
   where $F_{\rm Binom}(x; n, p)$ is the cumulative distribution function of the binomial distribution
   with parameters $n$ and $p$.
\end{proposition}

Using this result in combination with either
the Chernoff bound or
Hoeffding's inequality, we can derive the following two bounds that show the asymptotic behavior:
\begin{corollary}
   \label{cor:percentile-bound-as-chernoff-hoeffding}
   We have
   \begin{multline}
      \Pr\left(\hat x_p \geq x_p \right)
      \geq \\ 1 - \exp \left( -n \kldiv{\Bernoulli(p + \epsilon)}{\Bernoulli(p)} \right)
      \label{eq:percentile-chernoff-bound}
   \end{multline}
   where $\kldiv{\Bernoulli(p + \epsilon)}{\Bernoulli(p)}$ is the KL divergence
   between the Bernoulli distribution with parameter $p + \epsilon$
   and the Bernoulli distribution with parameter $p$.
   A simpler, but less tight lower bound is given by
   \begin{equation}
      \Pr\left(\hat x_p \geq x_p \right)
      \geq 1 - \exp \left( -2 n \epsilon^2 \right).
      \label{eq:percentile-hoeffding-bound}
   \end{equation}
\end{corollary}

We prove both results in \cref{sec:proof-percentile-bounds}.

We want to apply these results in the following situation.
We fix $0 < p, \delta < 1$, $\epsilon > 0$ with $p + \epsilon < 1$
and want to know how many points we need to sample such that we get
\begin{equation}
   \Pr(\hat x_p \geq x_p) \geq \delta.
   \label{eq:percentile-delta-definition}
\end{equation}
By setting the right hand sides of the equations above equal to $\delta$
and solving for $n$, we obtain the values of $n$ shown in
\cref{table:percentile-minimum-samples}.

In the results in \cref{sec:results-with-less-adversarial-patch-inputs},
we sample a million pairs of input and patch inputs independently.
Applying the results from this section,
if we take $\epsilon = \num{5e-4}$,
then for the 95th-percentile,
$\delta = 0.9891$,
whereas for the 99-th, 99.9-th and 99.99-th percentile,
$\delta$ is indistinguishably close to 1 in the \texttt{stats.scipy} package.
For our main results (described in \cref{sec:results}), however,
we independently sample a thousand inputs and a thousand patch inputs,
for a total of a one million pairs.
However, these million pairs are not independent, so the results
from this section do not apply directly.
We can instead consider them as 1000 i.i.d. samples,
even though that underestimates the true results,
and then some reasonable numbers to consider would be the following:
for $\epsilon = \num{0.01}$ and $p=0.95$,
we find $\delta = 0.9194$;
for $\epsilon = \num{5e-3}$ and $p = 0.99$,
we find $\delta = 0.9339$.

\begin{table}
   \small
   \caption{
      A table showing how many samples are needed such that $\hat x_p$
      is an upper bound of the $p$-th percentile of a real-valued distribution with probability
      at least $\delta$; i.e., such that we have $\Pr(\hat x_p \geq x_p) \geq \delta$ (equation \eqref{eq:percentile-delta-definition}).
      The column labeled ``exact'' uses the exact calculation from equation (equation \eqref{eq:percentile-binomial-calculation});
      the column ``Chernoff'' uses the formula based on the Chernoff bound (equation \eqref{eq:percentile-chernoff-bound});
      and the column ``Hoeffding'' uses the formula based on Hoeffding's inequality (equation \eqref{eq:percentile-hoeffding-bound}).
      The ``exact'' column will always provide a more precise value;
      the other columns are included to give the reader a sense of the approximation quality.
   }
\begin{tblr}{ccc|rrr}
\toprule
$p$ & $\delta$ & $\epsilon$ & \centering exact & \centering Chernoff & \centering Hoeffding \\
\midrule
  0.95 & 0.95 & 0.01       & \num{1282}    & \num{2659}   & \num{14979}      \\
  0.95 & 0.99 & 0.01       & \num{2437}    & \num{4088}   & \num{23026}      \\
  0.95 & 0.95 & 0.04       & \num{59}      & \num{122}    & \num{937}        \\
  0.99 & 0.95 & 0.005      & \num{1049}     & \num{1937}   & \num{59915}      \\
  0.99 & 0.99 & 0.005      & \num{1736}    & \num{2978}   & \num{92104}      \\
  0.999 & 0.999 & 0.0005   & \num{31236}   & \num{44987}  & \num{13815511}   \\
\bottomrule
\end{tblr}
\label{table:percentile-minimum-samples}
\end{table}

\section{Results}
\label{sec:results}

We applied the method described in \cref{sec:methodology} to three circuits found previously in the literature:
the circuits for the IOI task \cite{wangInterpretabilityWildCircuit2022},
for the docstring task \cite{heimersheimCircuitPythonDocstrings2023},
and for the greater-than task \cite{hannaHowDoesGPT22023}.
In each of these tasks, the model needs to complete a sentence created from a task-specific template.
For example, in the IOI task, the template is of the form
``Afterwards, \tempval{name1} and \tempval{name2} went to the \tempval{place}. \tempval{name2} gave a \tempval{object} to ....'',
and the model's task is to complete the sentence with the token for \tempval{name1}.
These task-specific templates are described in \cref{table:tasks}.

First, to determine the circuit's explanatory power over a range of inputs,
we sampled 1000 clean inputs and 1000 corrupted inputs, for a total of 1 million pairs for each task.
We used the same data distributions as in \cite{conmyAutomatedCircuitDiscovery2023};
these distributions sample each template value from a pre-determined list with equal probability.
A crucial difference, though, is that we mixed all clean inputs with all corrupted inputs,
whereas the original datasets paired them up in more restrictive ways.
For example, in IOI, we allowed the corrupted input point to involve a different object and place,
whereas originally the clean and corrupted inputs coincided on everything but the names.
In \cref{sec:results-with-less-adversarial-patch-inputs},
we argue for this approach over only using the corrupted inputs that were used in the original dataset,
but for completeness, that section also contains our evaluation of the circuit
using only the corrupted inputs from the original dataset.

\begin{figure}[h]
    \centering
    \includegraphics[width=0.4\textwidth]{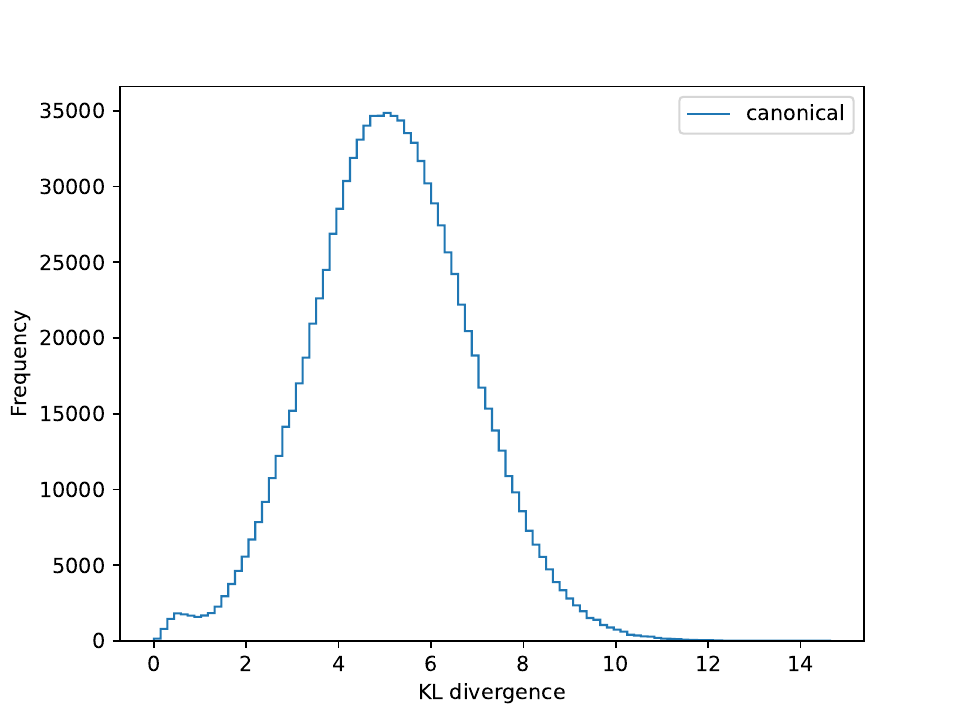} % Replace 'your_svg_file' with the actual filename without the .svg extension
    \caption{A histogram of the KL divergence for the IOI task.
    The x-axis shows the KL divergence between the model's output and the circuit's output on an input-corrupted-input pair,
    and the y-axis shows the number of input-corrupted-input pairs from our random sample of 1 million points
    that fall into each bin.
    There are 100 bins of equal size between the values of 0 and the maximum KL divergence achieved.
    Summary statistics of the plotted distribution are displayed in \cref{table:percentiles}.
   }
    \label{fig:ioi-histogram}
\end{figure}

\begin{figure}[h]
    \centering
    \includegraphics[width=0.4\textwidth]{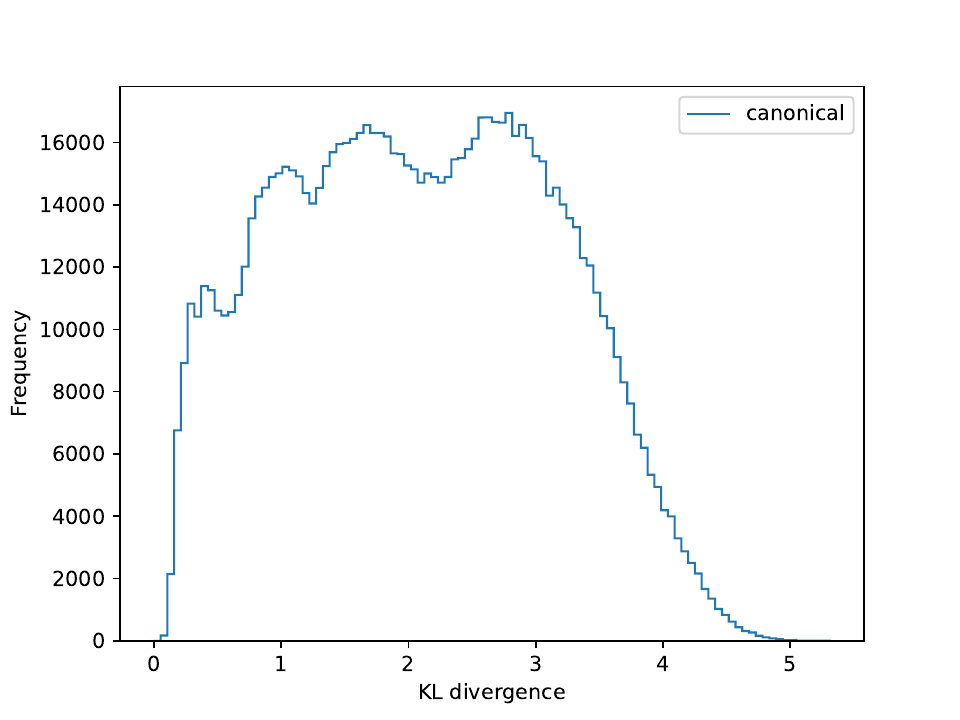} % Replace 'your_svg_file' with the actual filename without the .svg extension
    \caption{A histogram of the KL divergence for the greater-than task.}
    \label{fig:greaterthan-histogram}
\end{figure}

\begin{figure}[h]
    \centering
    \includegraphics[width=0.4\textwidth]{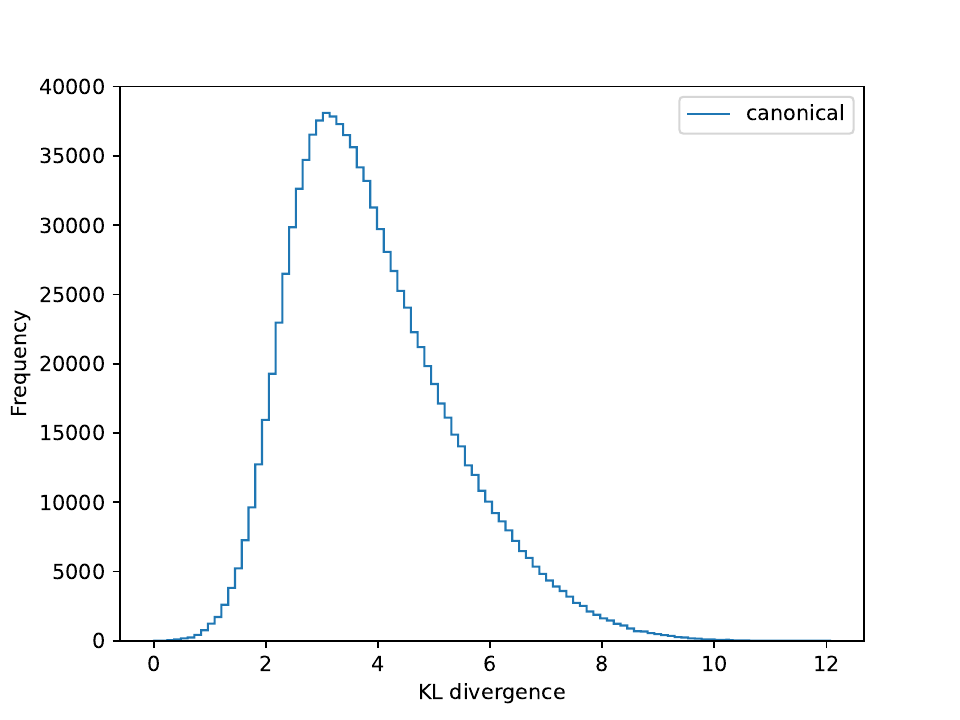} % Replace 'your_svg_file' with the actual filename without the .svg extension
    \caption{A histogram of the KL divergence for the docstring task.}
    \label{fig:docstring-histogram}
\end{figure}

We then calculated the KL divergence between the model's output and the circuit's output
for all those pairs and plotted the results as histograms in
\crefrange{fig:ioi-histogram}{fig:docstring-histogram}.
Summary statistics for these distributions are displayed in 
\cref{table:percentiles}.
The high percentiles and the max KL divergence shown in that table can then be considered as the adversarial evaluation metrics.

Secondly, to get a better understanding of the worst-case behavior of these circuits,
we took the top 10 worst-performing (input, corrupted input) pairs for each circuit,
and performed a forward-pass on the circuit and the model to obtain the top three most likely outputs
for both the model and the circuit.
These results are displayed in \cref{sec:tables-worst-performing-input-points},
\crefrange{table:ioi-outputs}{table:greaterthan-outputs}.
We discuss some of their implications below.

\begin{table*}
   \small
   \caption{A table summarizing the tasks for the circuits we analyze.
      Note that in each case, the template values in the corrupted input
      are completely independent from the template values in the clean input.}
   \begin{tblr}{c r X[l] X[l] c X[l]}
      \toprule
      Task && Input template & Expected output & LLM & Notes \\
      \midrule
      \SetCell[r=2]{m} Docstring
           & clean:
           & Python function definitions with a docstring that starts describing
           the function's parameters, but crucially does not list all parameters.
           & The name of the next undescribed parameter in the docstring.
           & \SetCell[r=2]{m} \texttt{attn-only-4l}\footnote{See \url{https://neelnanda-io.github.io/TransformerLens/generated/model_properties_table.html}}
           & \SetCell[r=2]{m}\phantom{none}
           \\
           % task (multi-row)
           & corrupted:
           & Similar to clean, but the parameters in the docstring are not
             necessarily the same as in the function definition.
           & % no expected output
           & % model (multi-row)
           & % notes (multi-row)
           \\
           \hline
      \SetCell[r=2]{m} Greater-than
           & clean:
           & ``The \tempval{noun} lasted from the year \tempval{year $d_1d_2d_3d_4$} to \tempval{$d_1d_2$}''
           & any 2-digit number higher than $d_3d_4$
           & \SetCell[r=2]{m} gpt2-small % I believe this is https://huggingface.co/openai-community/gpt2
           & \SetCell[r=2]{m}\phantom{none}
           \\
           % task (multi-row)
           & corrupted:
           & Similar to clean, but the last two digits of the year are always \texttt{01}.
           & % no expected output
           & % model (multi-row)
           & % notes (multi-row)
           \\
           \hline
      \SetCell[r=2]{m} IOI
           & clean:
           & ``Afterwards, \tempval{name1} and \tempval{name2} went to the \tempval{place}. \tempval{name2} gave a \tempval{object} to ''
           & \tempval{name1}
           &  \SetCell[r=2]{m} gpt2-small
           & \SetCell[r=2]{m} This template can easily be varied, e.g. by switching the order of the names or 
           changing some of the non-templated words, such as replacing ``went to'' with ``decided to go to''.
           \\
           % task (multi-row)
           & corrupted:
           & ``Afterwards, \tempval{name1'} and \tempval{name2'} went to the \tempval{place'}. \tempval{name3'} gave a \tempval{object'} to ''
           & % no name
           & % model (multi-row)
           & % notes (multi-row)
           \\
      \bottomrule
   \end{tblr}
   \label{table:tasks}
\end{table*}

\begin{table*}[]
   \centering
   \caption{Summary statistics from the KL divergence distributions plotted in \crefrange{fig:ioi-histogram}{fig:docstring-histogram}.
      The columns labelled ``abs'' show the absolute values of the KL divergence, whereas the columns labelled
      ``z-score'' show the difference between the percentile and the mean expressed as a multiple of the standard deviation.
   }
   \input{percentiles.tex}
   \label{table:percentiles}
\end{table*}

\paragraph{Comparing the KL divergence distributions}
The table of summary statistics (\cref{table:percentiles})
for the distribution of KL divergences for the three tasks, computed on our random sample of 1 million input-corrupted-input pairs,
shows that each circuit's worst-case performance is quite far from its mean performance.
For the IOI and docstring tasks,
the standard deviation is quite large,
the worst points we found are more than 5 standard deviations away from the mean,
and the z-scores indicate that the distributions have slightly thicker tails than the normal distribution.
All of this indicates that it is worthwhile to pay attention to the tails of the distribution
when evaluating the circuit's performance.

\paragraph{Docstring}
A notable feature of the 10 worst-performing input pairs for the docstring task
is that 7 out of the 10 have the same corrupted input (\texttt{def image(self, key, file, ...)}),
which heavily skews the output logits towards the parameters from that corrupted input (notably the \texttt{file} parameter).
This indicates that there are some components outside of the circuit that play a strong role in this task
and perhaps only activate on certain inputs.

\paragraph{Greater-than}
The greater-than circuit is the best-performing circuit of the three:
the worst-performing input pairs have a much lower KL divergence,
and their KL divergence does not deviate as much from the mean as the other two tasks
(see \cref{table:percentiles}).
The output analysis in \cref{table:greaterthan-outputs} shows that even in the 10 worst cases,
the circuit's most likely output always coincides with the model's most likely output,
and the top three most likely outputs are almost always admissible.

Moreover, unlike in the other two tasks,
there is a straightforward explanation that could have been predicted in advance:
the worst-performing points are those where the clean input
has a very high two-digit number (e.g., 94),
so that there are very few allowed completions (only the two-digit numbers $> 94$),
whereas the corrupted input has a very low two-digit number (e.g. 01),
allowing almost all two-digit numbers as completion.
If we assume that the full model's output distributions are approximately uniform over all allowable two-digits completions,
for example,
then the KL divergence between the clean input's output and the corrupted input's output is maximal,
which plausibly explains why these inputs are the worst-performing inputs.
We provide more evidence for this claim in \ref{sec:data-circuit-performance-by-field}.

\paragraph{IOI}
One striking feature of the worst-performing input pairs for the IOI task
is that they often seem to involve romantic items.
We provide more evidence for this observation in \ref{sec:data-circuit-performance-by-field}.
This behavior was even more apparent in earlier iterations of our experiment
where we fixed the corrupted input.
A plausible hypothesis is that
parts of the model outside of the circuit are dormant in normal contexts
but activate when romantic items are involved.

It is also worth noting that the IOI dataset from \cite{conmyAutomatedCircuitDiscovery2023} that we used
only has eight possible values
\footnote{The objects are: 
    \textsl{ring},
    \textsl{kiss},
    \textsl{bone},
    \textsl{basketball},
    \textsl{computer},
    \textsl{necklace},
    \textsl{drink},
    and
    \textsl{snack}.}
for the object being given.
It seems plausible that the circuit could behave very poorly on other objects as well.

\section{Discussion}
\label{sec:discussion-high-level}

We have found that the IOI and Docstring circuits can produce very different outputs
than the full model, even on inputs from the original task.
In both cases, the worst-case performance is quite far from the mean performance.
This casts doubt on the possibility of using these circuits to understand the full model's behavior.
We expect this discrepancy to be even worse on untested input data or under minor distributional shifts:
what happens when Mary \textsl{has secret plans} to give \textsl{an atomic bomb}?

Some of the badly performing inputs seem to follow a pattern,
e.g., IOI's failure in romantic contexts and Docstring's tendency to pick up on the \texttt{file} parameter in the corrupted input.
It seems likely that we could improve the circuits by addressing these specific issues.
However, there are also aspects of the circuits' failure that seem more random and inscrutable,
and it is unclear if these issues can be fixed, or if there is some fundamental,
inherent limitation to the circuits' explanatory power.

We conclude that it is important to find circuits that are more robust,
and speculate that we might achieve this by using adversarial evaluation metrics in circuit discovery techniques.

\section{Future Work}

This paper proposes a method for evaluating circuits adversarially.
As we have already alluded to, these evaluation criteria could be integrated into circuit discovery algorithms.
In future work, we aim to do this and test its effectiveness.
It might improve both the average and worst-case performance.

Additionally, the hope is that this will lead to circuits that are more robust under distributional shifts.
The results of this paper show that even under small changes in the input,
the circuit can lose its explanatory power.
If we want to use circuits in safety-critical applications,
they need to be more robust.
It would be worthwhile to measure how robust current circuits are to distributional
and to try to improve this robustness.

\section*{Acknowledgements}
We thank MATS, AI Safety Support, and the Long-Term Future Fund (LTFF) for providing funding for this research
and providing an amazing research environment.
Additional, we are grateful for the compute resources provided by FAR AI,
and for ChengCheng Tan for her editorial assistance in improving the readability of this paper.
We also thank Iv\'an Arcuschin Moreno for helpful discussions that improved this research,
and Thomas Kwa for working on the edge-level subnetwork probing code that we adapted
to run forward passes with ablations.

\section*{Impact statement}
This paper aims to advance the field
of mechanistic interpretability. While there are many potential societal
consequences of our work, none need to be
specifically highlighted here.

\bibliography{adv_opt}
\bibliographystyle{icml2024}

\newpage
\appendix
\onecolumn % optional

\section{Tables of worst-performing input points}
\label{sec:tables-worst-performing-input-points}

See \crefrange{table:ioi-outputs}{table:greaterthan-outputs} for the tables of (input, corrupted input) pairs
on which the circuits perform the worst,
together with the most likely outputs for those inputs.
See \cref{sec:results} for more details.

% Tables generated with acdc/nudb/adv_opt/analysis/a2024_w21_01_generate_latex_tables_with_circuit_outputs.py
\begin{table}[H]
   \small \centering
   \caption{Top 10 worst-performing input pairs and corresponding 3 most likely outputs for the IOI task.
      The first two columns show the top 10 worst-performing input pairs, with the worst on top.
      The third column displays the KL divergence between the full model's output,
      and the circuit's output when run with resample ablation using the patch input,
      as explained in detail in \cref{sec:methodology}.
      The last 6 columns show the three most likely output tokens for the model and the circuit,
      with that output's unnormalized logit score shown in parentheses beneath it.
   }
   \input{ioi-outputtable.tex}
   \label{table:ioi-outputs}
\end{table}

\begin{table}[H]
   \tiny \centering
   \caption{Top 10 worst-performing input pairs and corresponding 3 most likely outputs for the docstring task.}
   \input{docstring-outputtable.tex}
   \label{table:docstring-outputs}
\end{table}

\begin{table}[H]
   \small \centering
   \caption{Top 10 worst-performing input pairs and corresponding 3 most likely outputs for the greater-than task.}
   \input{greaterthan-outputtable.tex}
   \label{table:greaterthan-outputs}
\end{table}

\section{Results with less adversarial patch inputs}
\label{sec:results-with-less-adversarial-patch-inputs}

For our main results in \cref{sec:results}, we took a very adversarial approach towards
the patch inputs:
any clean input could be paired with any patch input for resample ablation.
However, the circuits were originally found and tested with resample ablations
that were more restrictive:

\begin{itemize}
   \item In the IOI task, the location and object in the corrupted input were the same as in the clean input.
   \item In the docstring task, the only difference between the corrupted input and the clean input was the parameter names in the docstring.
   \item In the greater-than task, the event and the first two digits of the years in the corrupted input were the same as in the clean input.
\end{itemize}

(We recall that a short description of these tasks is shown in \cref{table:tasks}.)

We believe the our more adversarial approach that allows any corrupted input to be matched with any clean input,
is justified for the following reasons:

\begin{itemize}
   \item The high-level explanation in \cite{wangInterpretabilityWildCircuit2022}
      suggests that the model identifies the indirect object through a mechanism that
      does not depend in any way on the location or object.
      Some attention heads are name identifiers, others duplicate detectors, others name inhibitors -- none of these depend on the location or object.
   \item The additional information that is ablated is not necessary to complete the task.
   \item The tasks inputs still follow the same restrictive template.
\end{itemize}

For completeness, this section presents the adversarial evaluation metrics
on random samples of 1 million input-corrupted-input pairs
where the patch inputs are matched in the same way as in the original dataset.

\Cref{table:percentiles-matched} shows
the standard deviations and the means are lower than if we allow any corrupt input,
indicating that the circuits indeed perform better on these matched input-corrupted-input pairs.
However, the worst points are many standard deviations (9.97 and 15.47 for docstring and IOI, respectively)
removed from the mean,
so there are still inputs on which the circuits perform very poorly.

The table with the top 10 worst inputs for IOI (\cref{table:ioi-outputs-matched})
shows that many of our conclusions still hold:
the worst inputs look very benign, but the model correctly predicts the next token,
whereas the circuit either takes a name from the patched input or repeats the subject rather than identifying the indirect object,
with very high confidence.

The greater-than circuit performs very well, and on the top 10 worst inputs, all the most likely tokens are permissible.
Most of the 10 worst inputs for the docstring task usually predict a token that is indeed one of the parameters,
but it has already occurred before in the clean input.

\begin{figure}[H]
    \centering
    \includegraphics[width=0.4\textwidth]{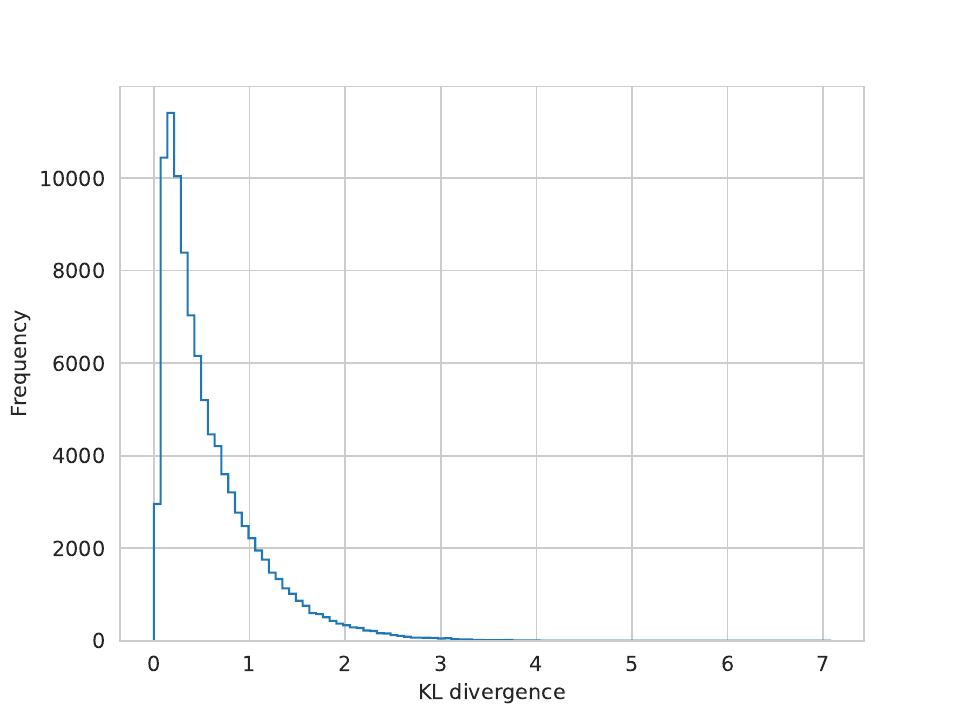} % Replace 'your_svg_file' with the actual filename without the .svg extension
    \caption{A histogram of the KL divergence for the IOI task,
       where all input-corruputed-input pairs are matched in the same way as in the original dataset,
       i.e., with the same location and object in the corrupted input as in the clean input.
    The x-axis shows the KL divergence between the model's output and the circuit's output on an input-corrupted-input pair,
    and the y-axis shows the number of input-corrupted-input pairs from our random sample of 1 million points
    that fall into each bin.
    There are 100 bins of equal size between the values of 0 and the maximum KL divergence achieved.
    Summary statistics of the plotted distribution are displayed in \cref{table:percentiles}.
   }
    \label{fig:ioi-histogram-matched}
\end{figure}

\begin{figure}[H]
    \centering
    \includegraphics[width=0.4\textwidth]{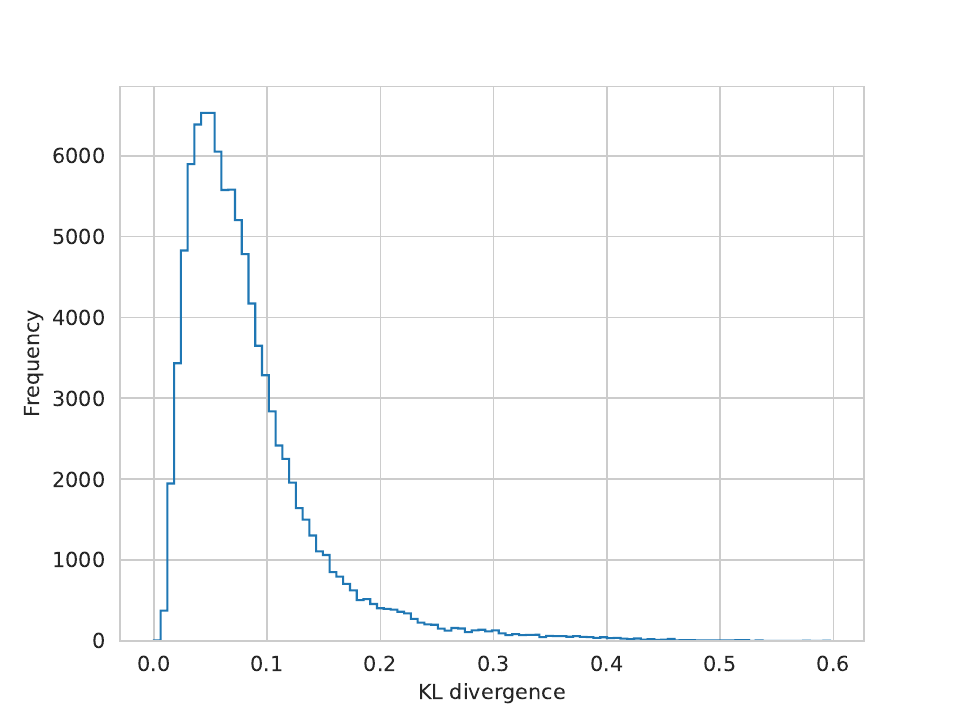} % Replace 'your_svg_file' with the actual filename without the .svg extension
    \caption{A histogram of the KL divergence for the greater-than task
       where all input-corruputed-input pairs are matched in the same way as in the original dataset,
       i.e., with the same event and first two digits in the corrupted input as in the clean input.
    }
    \label{fig:greaterthan-histogram-matched}
\end{figure}

\begin{figure}[H]
    \centering
    \includegraphics[width=0.4\textwidth]{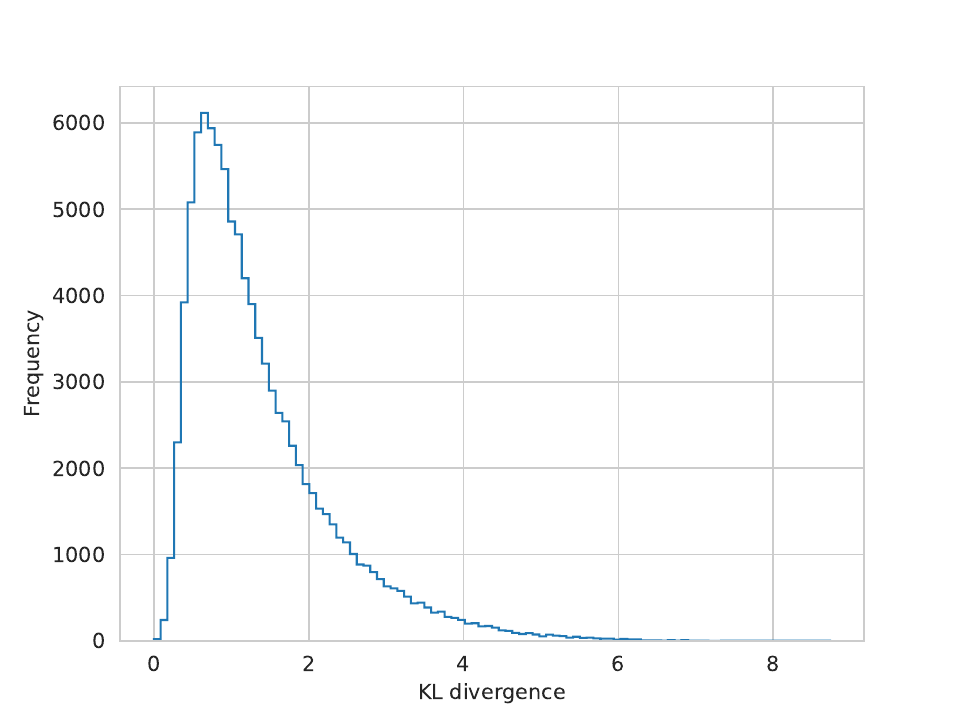} % Replace 'your_svg_file' with the actual filename without the .svg extension
    \caption{A histogram of the KL divergence for the docstring task
       where all input-corruputed-input pairs are matched in the same way as in the original dataset,
       i.e., the only difference between the corrupted input and the clean input is the parameter names
    in the docstring (but the parameters in the function signature are the same).}
    \label{fig:docstring-histogram-matched}
\end{figure}

\begin{table}[H]
   \centering
   \caption{Summary statistics from the KL divergence distributions plotted in \crefrange{fig:ioi-histogram-matched}{fig:docstring-histogram-matched}.
      The columns labelled ``abs'' show the absolute values of the KL divergence, whereas the columns labelled
      ``z-score'' show the difference between the percentile and the mean expressed as a multiple of the standard deviation.
   }
   \input{percentiles-matched.tex}
   \label{table:percentiles-matched}
\end{table}

% Tables generated with acdc/nudb/adv_opt/analysis/a2024_w21_01_generate_latex_tables_with_circuit_outputs.py
\begin{table}[H]
   \small \centering
   \caption{Top 10 worst-performing input pairs and corresponding 3 most likely outputs for the IOI task
      where all input-corrupted-input pairs are matched in the same way as in the original dataset.
      The first two columns show the top 10 worst-performing input pairs, with the worst on top.
      The third column displays the KL divergence between the full model's output,
      and the circuit's output when run with resample ablation using the patch input,
      as explained in detail in \cref{sec:methodology}.
      The last 6 columns show the three most likely output tokens for the model and the circuit,
      with that output's unnormalized logit score shown in parentheses beneath it.
   }
   \input{ioi-outputtable-matched-corruptions.tex}
   \label{table:ioi-outputs-matched}
\end{table}

\begin{table}[H]
   \tiny \centering
   \caption{Top 10 worst-performing input pairs and corresponding 3 most likely outputs for the docstring task
   where all input-corrupted-input pairs are matched in the same way as in the original dataset.}
   \input{docstring-outputtable-matched-corruptions.tex}
   \label{table:docstring-outputs-matched}
\end{table}

\begin{table}[H]
   \small \centering
   \caption{Top 10 worst-performing input pairs and corresponding 3 most likely outputs for the greater-than task
   where all input-corrupted-input pairs are matched in the same way as in the original dataset.}
   \input{greaterthan-outputtable-matched-corruptions.tex}
   \label{table:greaterthan-outputs-matched}
\end{table}

\section{Proof of Percentile Bounds}
\label{sec:proof-percentile-bounds}

In this appendix, we prove
\cref{thm:percentile-bound-as-binomial} and \cref{cor:percentile-bound-as-chernoff-hoeffding}.

We remind the reader of the setup.
Let $X$ be some randomly distributed variable,
let $0 < p < 1$, and let $\epsilon > 0$ with $p + \epsilon < 1$.
Suppose we have a sample of $n$ i.i.d. draws from $X$
and we want to use that sample to find an upper bound
of the real (but unknown) $p$-th percentile of $X$.
We denote the real $p$-th percentile by $x_p$
and we take as our estimate for the upper bound
\begin{equation}
  \hat x_p := \text{the $\ceil{n(p + \epsilon)}$-th element in the sample ordered by value from low to high.}
  \label{eq:definition-hat-xp}
\end{equation}
Note that $\epsilon$ can be considered a kind of safety margin:
by making it bigger, we get a less tight estimate of the upper bound,
but we increase the probability that it is actually an upper bound.

\begin{proof}[Proof of \cref{thm:percentile-bound-as-binomial}]
  The probability $\Prob(\hat x_p \geq x_p)$ is the same as the probability
  that fewer than $\ceil{n(p + \epsilon)}$ elements from our sample come from the lower
  $p$ percentiles of the distribution --- indeed, this is equivalent to saying
  that the $\ceil{n(p + \epsilon)}$-th element comes from the upper $1 - p$ percentiles,
  and is hence at at least as large as $x_p$.

  We can calculate this probability with the binomial distribution $\Binom(n, p)$,
  because the probability of drawing a sample from the lower $p$ percentiles is precisely $p$.
\end{proof}

Let $Y \sim \Binom(n, p)$ be a binomially distributed random variable,
and let $p < a < 1$.
Then the Chernoff bound \citep[Theorem 1]{Arratia1989TutorialOL}
says
\begin{equation}
   \Pr(Y \geq an) \leq \exp(-n \kldiv{\Bernoulli(a)}{\Bernoulli(p)}).
   \label{eq:chernoff-bound}
\end{equation}
Alternatively, Hoeffding's inequality \citep[Theorem 1]{Hoeffding1963ProbabilityIO} says
\begin{equation}
   \Pr\left(\frac 1 n Y - p \geq a - p\right) \leq \exp(-n (a - p)^2)
\end{equation}
which we can rewrite to
\begin{equation}
   \Pr(Y \geq an) \leq \exp(-n (a - p)^2)
   \label{eq:hoeffding-inequality}.
\end{equation}

\begin{proof}[Proof of \cref{cor:percentile-bound-as-chernoff-hoeffding}]
   \Cref{thm:percentile-bound-as-binomial} tells us
   \begin{equation}
      \Pr\left(\hat x_p \geq x_p \right) = F_{\rm Binom}(\lceil (p + \epsilon) \cdot n \rceil - 1; n, p)
   \end{equation}
   We can rewrite the right hand side:
   \begin{equation}
      \begin{split}
         F_{\rm Binom}(\lceil (p + \epsilon) \cdot n \rceil - 1; n, p)
         &= \Pr\left(Y \leq \lceil (p + \epsilon) \cdot n \rceil - 1 \right) \\
         &= 1 - \Pr\left(Y \geq \lceil (p + \epsilon) \cdot n \rceil \right)
      \end{split}
   \end{equation}
   where $Y \sim \Binom(n, p)$.
   Applying the Chernoff bound \eqref{eq:chernoff-bound},
   or alternatively applying the Hoeffding inequality \eqref{eq:hoeffding-inequality},
   gives us the inequalities we're looking for.
\end{proof}

\section{Analysis of circuit performance grouped by prompt fields}
\label{sec:data-circuit-performance-by-field}

In \ref{sec:results},
we remarked on some patterns in the top 10 worst-performing inputs
listed in \cref{sec:tables-worst-performing-input-points}.
In this appendix, we provide additional support for those claims,
by not just looking at the top 10 worst-performing inputs,
but by grouping all inputs based on a template value in their prompt
(e.g. in the IOI task, the place, or the object that is being given).
The data shows that certain template values lead to higher losses more often.

\begin{figure}[H]
    \centering
    \includegraphics[width=1.0\textwidth]{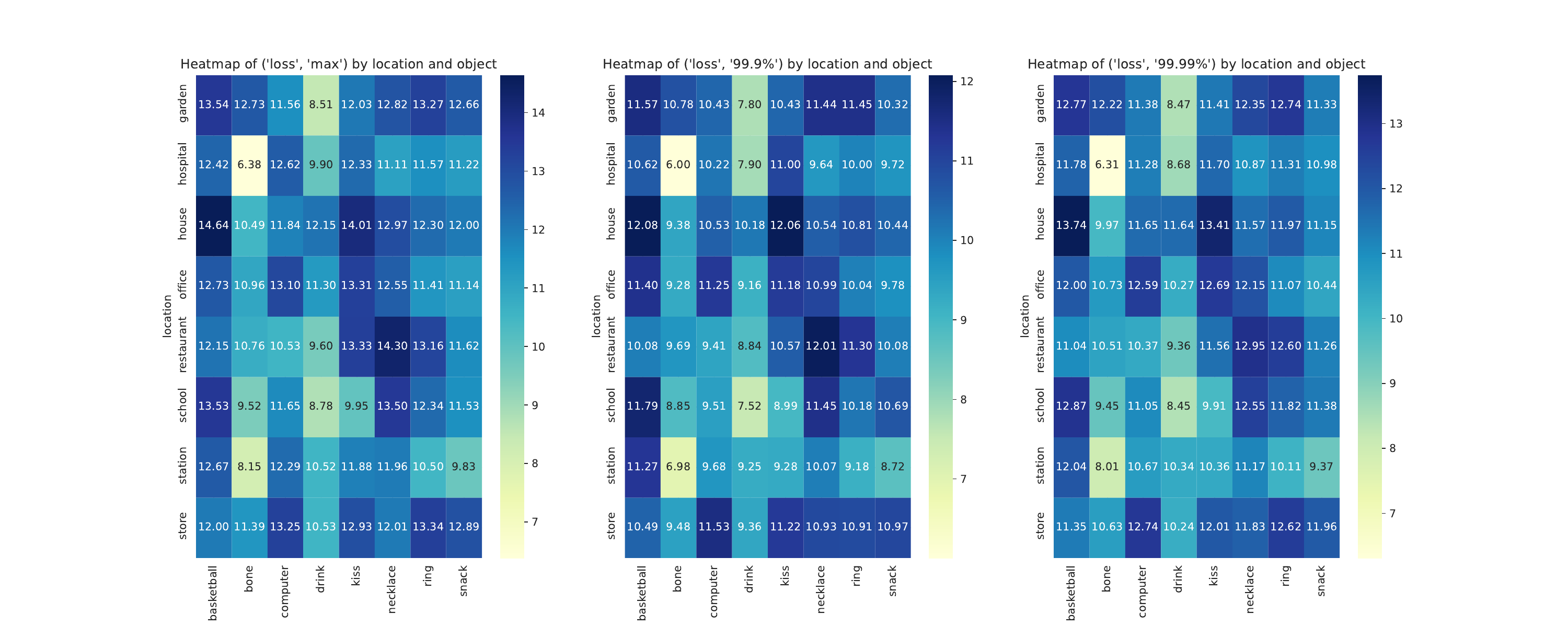} % Replace 'your_svg_file' with the actual filename without the .svg extension
    \caption{
       IOI: Three heatmaps, showing different percentiles (the max, the 99.9th percentile, and the 99.99th percentile)
       of the distribution of KL divergences between the circuit's output and the model's output on a sample of 1 million input-corrupted-input pairs,
       as in \cref{sec:results},
       plotted against the location and the object in the clean input prompt.
   }
    \label{fig:ioi-heatmap-loss-distribution}
\end{figure}

\begin{figure}[H]
    \centering
    \includegraphics[width=1.0\textwidth]{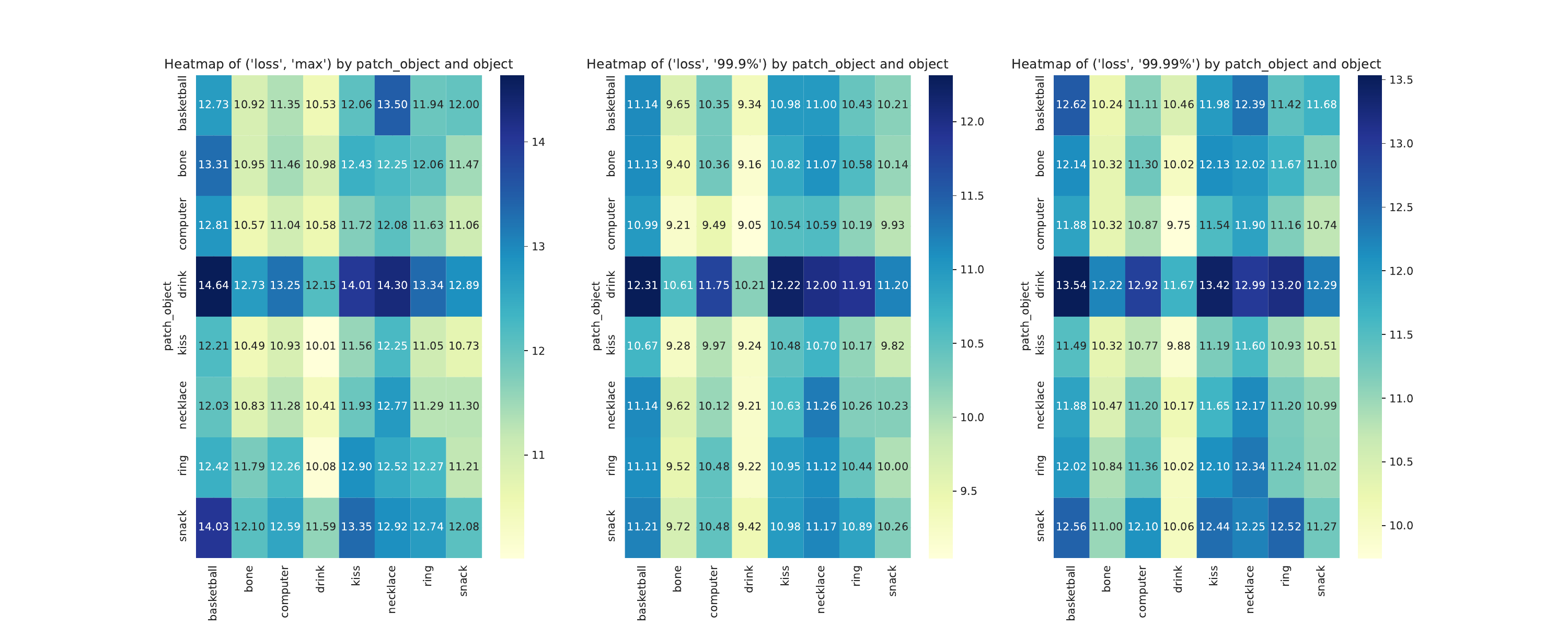} % Replace 'your_svg_file' with the actual filename without the .svg extension
    \caption{
       IOI: Three heatmaps, showing different percentiles (the max, the 99.9th percentile, and the 99.99th percentile)
       of the distribution of KL divergences between the circuit's output and the model's output on a sample of 1 million input-corrupted-input pairs,
       as in \cref{sec:results},
       plotted against the object in the clean input and the object in the patch input.
   }
    \label{fig:ioi-heatmap-loss-distribution-patch-object}
\end{figure}

\begin{figure}[H]
    \centering
    \includegraphics[width=1.0\textwidth]{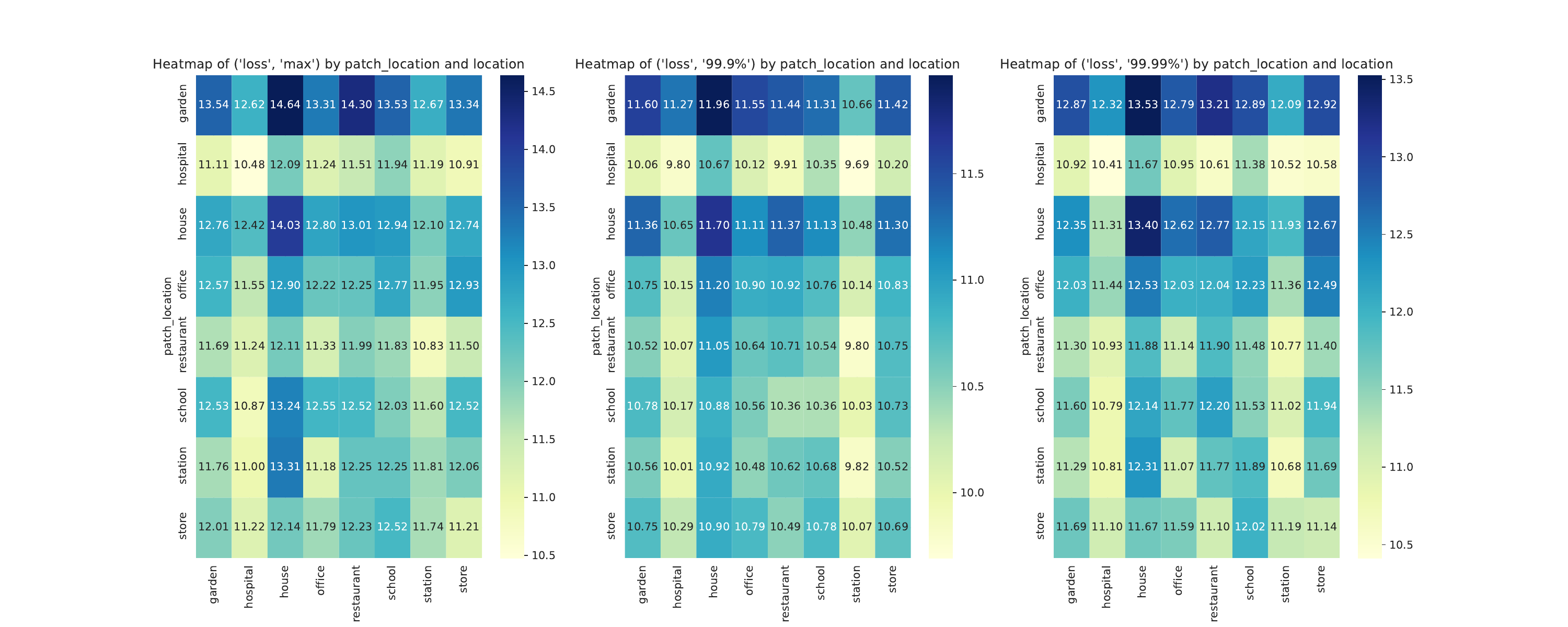} % Replace 'your_svg_file' with the actual filename without the .svg extension
    \caption{
       IOI: Three heatmaps, showing different percentiles (the max, the 99.9th percentile, and the 99.99th percentile)
       of the distribution of KL divergences between the circuit's output and the model's output on a sample of 1 million input-corrupted-input pairs,
       as in \cref{sec:results},
       plotted against the location in the clean input and the location in the patch input.
   }
    \label{fig:ioi-heatmap-loss-distribution-patch-location}
\end{figure}

For IOI,
\crefrange{fig:ioi-heatmap-loss-distribution}{fig:ioi-heatmap-loss-distribution-patch-location}
show that the performance of the IOI circuit in the higher percentiles varies considerably
with the object and the location that appear in the clean input prompt.
The more romantic objects, such as ``kiss'' and ``necklace'', perform especially poorly,
but there are also other objects and object-location combinations that perform poorly.
In future work we hope to find a mechanistic explanation for the circuit's failure in these cases.

For the docstring task,
we could not identify and then statistically confirm a clear hypothesis
for why some inputs fared much worse than others.

\begin{figure}[H]
    \centering
    \includegraphics[width=1.0\textwidth]{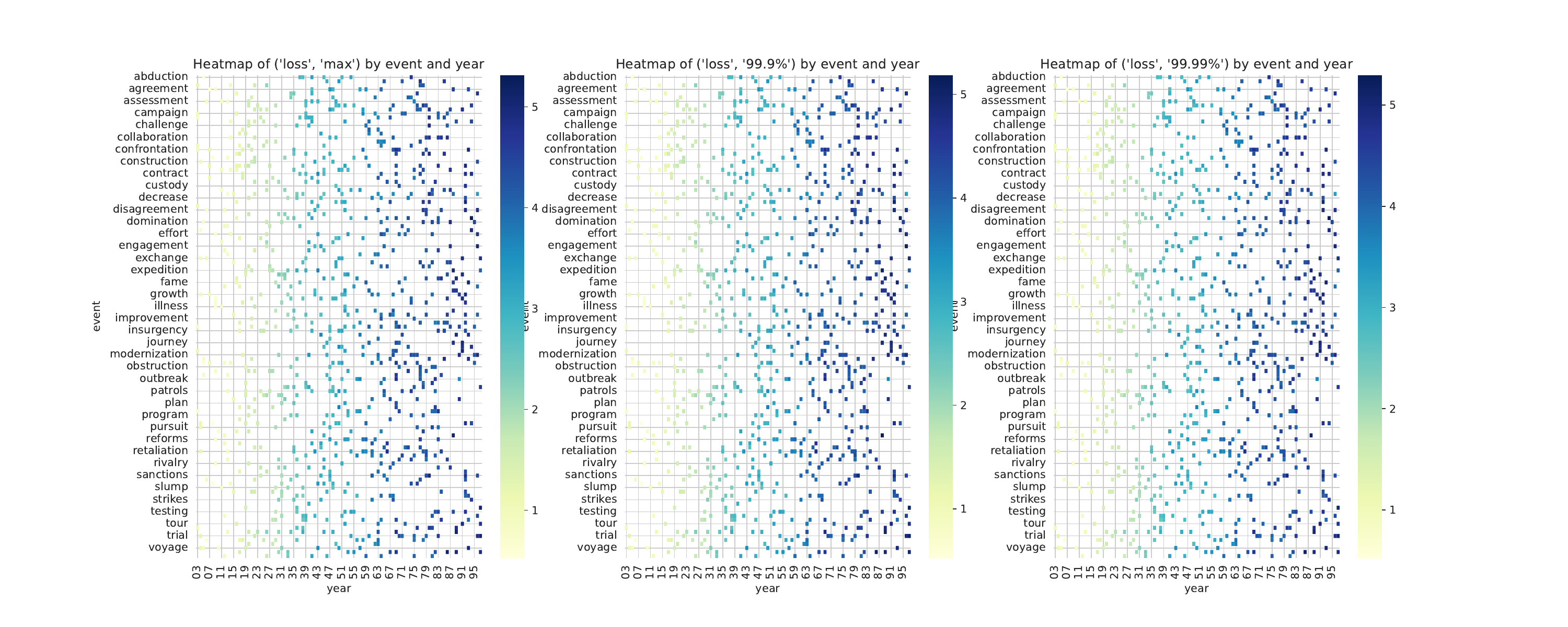} % Replace 'your_svg_file' with the actual filename without the .svg extension
    \caption{
       Greater-Than: Three heatmaps, showing different percentiles (the max, the 99.9th percentile, and the 99.99th percentile)
       of the distribution of KL divergences between the circuit's output and the model's output on a sample of 1 million input-corrupted-input pairs,
       as in \cref{sec:results},
       plotted against the event and the year in the clean input prompt.
   }
    \label{fig:greaterthan-heatmap-loss-distribution}
\end{figure}

For the greater-than task,
\cref{fig:greaterthan-heatmap-loss-distribution}
confirms that the circuit performs especially well when the last two digits of the year in the clean input are very low (e.g. 1705),
and especially poorly when the last two digits are very high (e.g. 1789),
as remarked towards the end of \cref{sec:results}.

\end{document}

%% file: percentiles.tex
\begin{tblr}{lrrrrrr}
\toprule
   & \SetCell[c=2]{c} docstring  & & \SetCell[c=2]{c} greaterthan & & \SetCell[c=2]{c}{ioi} \\
 & \centering abs & \centering z-score & \centering abs & \centering z-score & \centering abs & \centering z-score \\
\midrule
count & 1000000.00 &  & 1000000.00 &  & 1000000.00 &  \\
mean & 3.91 &  & 2.09 &  & 5.15 &  \\
std & 1.45 &  & 1.04 &  & 1.70 &  \\
min & 0.10 & -2.63 & 0.08 & -1.92 & 0.03 & -3.01 \\
25\% & 2.85 & -0.73 & 1.23 & -0.82 & 4.01 & -0.67 \\
50\% & 3.66 & -0.17 & 2.07 & -0.01 & 5.12 & -0.02 \\
75\% & 4.75 & 0.58 & 2.91 & 0.79 & 6.27 & 0.66 \\
95\% & 6.66 & 1.90 & 3.77 & 1.61 & 7.99 & 1.67 \\
99\% & 8.03 & 2.85 & 4.23 & 2.05 & 9.25 & 2.41 \\
99.9\% & 9.46 & 3.84 & 4.63 & 2.44 & 10.81 & 3.33 \\
99.99\% & 10.58 & 4.61 & 4.91 & 2.71 & 12.25 & 4.17 \\
max & 12.07 & 5.64 & 5.31 & 3.09 & 14.64 & 5.57 \\
\bottomrule
\end{tblr}

%% file: ioi-outputtable.tex
\begin{tblr}{X[l] X[l] c | c c c | c c c}
\toprule
      & & & \SetCell[c=3]{c} model &&& \SetCell[c=3]{c} circuit && \\
   input & patch input & loss & 1st   & 2nd & 3rd & 1st     & 2nd & 3rd \\
\midrule
Then, Tiffany and Sean went to the house. Sean gave a basketball to & Then, Samuel and Adam went to the garden. Daniel gave a drink to & 14.64 & \shortstack{' Tiffany'\\ (19.60)} & \shortstack{' the'\\ (14.24)} & \shortstack{' Sean'\\ (13.91)} & \shortstack{' them'\\ (17.16)} & \shortstack{' the'\\ (16.74)} & \shortstack{' Daniel'\\ (15.45)} \\
Then, Crystal and Tyler went to the restaurant. Tyler gave a necklace to & Then, Samuel and Adam went to the garden. Daniel gave a drink to & 14.30 & \shortstack{' Crystal'\\ (18.17)} & \shortstack{' the'\\ (14.59)} & \shortstack{' a'\\ (13.30)} & \shortstack{' them'\\ (16.80)} & \shortstack{' the'\\ (16.56)} & \shortstack{' Samuel'\\ (15.68)} \\
Then, Tiffany and Sean went to the house. Sean gave a basketball to & Then, Samuel and Gregory went to the house. William gave a snack to & 14.03 & \shortstack{' Tiffany'\\ (19.60)} & \shortstack{' the'\\ (14.24)} & \shortstack{' Sean'\\ (13.91)} & \shortstack{' them'\\ (16.64)} & \shortstack{' the'\\ (16.31)} & \shortstack{' Tiffany'\\ (14.82)} \\
Then, Erica and Justin went to the house. Justin gave a kiss to & Then, Mark and David went to the garden. Paul gave a drink to & 14.01 & \shortstack{' Erica'\\ (19.99)} & \shortstack{' her'\\ (15.19)} & \shortstack{' the'\\ (15.07)} & \shortstack{' them'\\ (17.55)} & \shortstack{' the'\\ (16.41)} & \shortstack{' David'\\ (14.69)} \\
Then, Brittany and Brian went to the garden. Brian gave a basketball to & Then, Samuel and Adam went to the garden. Daniel gave a drink to & 13.54 & \shortstack{' Brittany'\\ (18.97)} & \shortstack{' Brian'\\ (15.04)} & \shortstack{' the'\\ (14.80)} & \shortstack{' them'\\ (17.16)} & \shortstack{' the'\\ (16.63)} & \shortstack{' Samuel'\\ (14.91)} \\
Then, Tiffany and Jason went to the school. Jason gave a basketball to & Then, Samuel and Adam went to the garden. Daniel gave a drink to & 13.53 & \shortstack{' Tiffany'\\ (18.30)} & \shortstack{' the'\\ (14.47)} & \shortstack{' her'\\ (13.88)} & \shortstack{' them'\\ (17.29)} & \shortstack{' the'\\ (16.75)} & \shortstack{' his'\\ (14.82)} \\
Then, Allison and Kevin went to the school. Kevin gave a necklace to & Then, Joseph and Joseph went to the garden. Thomas gave a basketball to & 13.50 & \shortstack{' Allison'\\ (19.08)} & \shortstack{' the'\\ (14.58)} & \shortstack{' her'\\ (14.08)} & \shortstack{' Allison'\\ (17.10)} & \shortstack{' them'\\ (15.63)} & \shortstack{' the'\\ (15.48)} \\
Then, Erica and Justin went to the house. Justin gave a kiss to & Then, Timothy and Samuel went to the house. Jesse gave a drink to & 13.47 & \shortstack{' Erica'\\ (19.99)} & \shortstack{' her'\\ (15.19)} & \shortstack{' the'\\ (15.07)} & \shortstack{' them'\\ (17.35)} & \shortstack{' the'\\ (16.56)} & \shortstack{' Timothy'\\ (15.14)} \\
Then, Erica and Justin went to the house. Justin gave a kiss to & Then, Samuel and Adam went to the garden. Daniel gave a drink to & 13.40 & \shortstack{' Erica'\\ (19.99)} & \shortstack{' her'\\ (15.19)} & \shortstack{' the'\\ (15.07)} & \shortstack{' them'\\ (17.01)} & \shortstack{' the'\\ (16.60)} & \shortstack{' his'\\ (14.97)} \\
Then, Erica and Justin went to the house. Justin gave a kiss to & Then, Benjamin and John went to the house. Charles gave a snack to & 13.35 & \shortstack{' Erica'\\ (19.99)} & \shortstack{' her'\\ (15.19)} & \shortstack{' the'\\ (15.07)} & \shortstack{' Erica'\\ (17.19)} & \shortstack{' the'\\ (15.93)} & \shortstack{' them'\\ (15.88)} \\
\bottomrule
\end{tblr}

%% file: docstring-outputtable.tex
\begin{tblr}{X[l] X[l] c | c c c | c c c}
\toprule
      & & & \SetCell[c=3]{c} model &&& \SetCell[c=3]{c} circuit && \\
   input & patch input & loss & 1st   & 2nd & 3rd & 1st     & 2nd & 3rd \\
\midrule
\texttt{\bossymbol \newlinesymbol
def port(self, order, match, fields, model, old, parent): \newlinesymbol
    """agent rule manager \newlinesymbol
 \newlinesymbol
    :param fields: set song \newlinesymbol
    :param model: plane action \newlinesymbol
    :param} & \texttt{\bossymbol \newlinesymbol
def image(self, key, file, filename, files, line, expected): \newlinesymbol
    """package crime framework \newlinesymbol
 \newlinesymbol
    :param host: dollar author \newlinesymbol
    :param command: cup spring \newlinesymbol
    :param} & 12.07 & \shortstack{' old'\\ (19.66)} & \shortstack{' new'\\ (17.75)} & \shortstack{' fields'\\ (16.64)} & \shortstack{' file'\\ (18.87)} & \shortstack{' filename'\\ (17.41)} & \shortstack{' line'\\ (16.74)} \\
\texttt{\bossymbol \newlinesymbol
def default(self, node, user, current, text, port, item): \newlinesymbol
    """export manager mission \newlinesymbol
 \newlinesymbol
    :param current: song spot \newlinesymbol
    :param text: delay draft \newlinesymbol
    :param} & \texttt{\bossymbol \newlinesymbol
def model(self, shape, message, group, file, result, fields): \newlinesymbol
    """content host bed \newlinesymbol
 \newlinesymbol
    :param new: share stage \newlinesymbol
    :param page: lift range \newlinesymbol
    :param} & 12.07 & \shortstack{' port'\\ (20.52)} & \shortstack{' current'\\ (15.69)} & \shortstack{' str'\\ (15.31)} & \shortstack{' port'\\ (17.82)} & \shortstack{' filename'\\ (17.39)} & \shortstack{' message'\\ (17.22)} \\
\texttt{\bossymbol \newlinesymbol
def default(self, node, user, current, text, port, item): \newlinesymbol
    """export manager mission \newlinesymbol
 \newlinesymbol
    :param current: song spot \newlinesymbol
    :param text: delay draft \newlinesymbol
    :param} & \texttt{\bossymbol \newlinesymbol
def image(self, key, file, filename, files, line, expected): \newlinesymbol
    """package crime framework \newlinesymbol
 \newlinesymbol
    :param host: dollar author \newlinesymbol
    :param command: cup spring \newlinesymbol
    :param} & 12.06 & \shortstack{' port'\\ (20.52)} & \shortstack{' current'\\ (15.69)} & \shortstack{' str'\\ (15.31)} & \shortstack{' file'\\ (19.44)} & \shortstack{' filename'\\ (17.97)} & \shortstack{' line'\\ (17.93)} \\
\texttt{\bossymbol \newlinesymbol
def create(self, token, field, request, content, order, new): \newlinesymbol
    """tree cut hell \newlinesymbol
 \newlinesymbol
    :param request: king bar \newlinesymbol
    :param content: income creation \newlinesymbol
    :param} & \texttt{\bossymbol \newlinesymbol
def image(self, key, file, filename, files, line, expected): \newlinesymbol
    """package crime framework \newlinesymbol
 \newlinesymbol
    :param host: dollar author \newlinesymbol
    :param command: cup spring \newlinesymbol
    :param} & 11.95 & \shortstack{' order'\\ (20.48)} & \shortstack{' request'\\ (18.78)} & \shortstack{' field'\\ (16.64)} & \shortstack{' file'\\ (20.19)} & \shortstack{' filename'\\ (17.68)} & \shortstack{' line'\\ (17.47)} \\
\texttt{\bossymbol \newlinesymbol
def values(self, json, module, count, end, model, index): \newlinesymbol
    """lead respect dust \newlinesymbol
 \newlinesymbol
    :param count: hell step \newlinesymbol
    :param end: volume pair \newlinesymbol
    :param} & \texttt{\bossymbol \newlinesymbol
def image(self, key, file, filename, files, line, expected): \newlinesymbol
    """package crime framework \newlinesymbol
 \newlinesymbol
    :param host: dollar author \newlinesymbol
    :param command: cup spring \newlinesymbol
    :param} & 11.90 & \shortstack{' model'\\ (21.46)} & \shortstack{' models'\\ (16.23)} & \shortstack{' id'\\ (15.41)} & \shortstack{' file'\\ (20.41)} & \shortstack{' filename'\\ (18.51)} & \shortstack{' line'\\ (17.95)} \\
\texttt{\bossymbol \newlinesymbol
def match(self, results, default, order, check, row, field): \newlinesymbol
    """activity path strength \newlinesymbol
 \newlinesymbol
    :param order: product plane \newlinesymbol
    :param check: fan bell \newlinesymbol
    :param} & \texttt{\bossymbol \newlinesymbol
def image(self, key, file, filename, files, line, expected): \newlinesymbol
    """package crime framework \newlinesymbol
 \newlinesymbol
    :param host: dollar author \newlinesymbol
    :param command: cup spring \newlinesymbol
    :param} & 11.88 & \shortstack{' row'\\ (20.65)} & \shortstack{' check'\\ (16.86)} & \shortstack{' bool'\\ (16.57)} & \shortstack{' file'\\ (19.56)} & \shortstack{' check'\\ (19.51)} & \shortstack{' line'\\ (18.09)} \\
\texttt{\bossymbol \newlinesymbol
def command(self, code, instance, create, size, sub, run): \newlinesymbol
    """border horse trip \newlinesymbol
 \newlinesymbol
    :param create: bishop attack \newlinesymbol
    :param size: duty horse \newlinesymbol
    :param} & \texttt{\bossymbol \newlinesymbol
def image(self, key, file, filename, files, line, expected): \newlinesymbol
    """package crime framework \newlinesymbol
 \newlinesymbol
    :param host: dollar author \newlinesymbol
    :param command: cup spring \newlinesymbol
    :param} & 11.80 & \shortstack{' sub'\\ (20.32)} & \shortstack{' run'\\ (15.97)} & \shortstack{' name'\\ (15.80)} & \shortstack{' file'\\ (20.00)} & \shortstack{' bool'\\ (17.52)} & \shortstack{' filename'\\ (17.31)} \\
\texttt{\bossymbol \newlinesymbol
def default(self, node, user, current, text, port, item): \newlinesymbol
    """export manager mission \newlinesymbol
 \newlinesymbol
    :param current: song spot \newlinesymbol
    :param text: delay draft \newlinesymbol
    :param} & \texttt{\bossymbol \newlinesymbol
def error(self, order, shape, match, filename, message, results): \newlinesymbol
    """star opening risk \newlinesymbol
 \newlinesymbol
    :param file: cycle second \newlinesymbol
    :param content: race staff \newlinesymbol
    :param} & 11.53 & \shortstack{' port'\\ (20.52)} & \shortstack{' current'\\ (15.69)} & \shortstack{' str'\\ (15.31)} & \shortstack{' item'\\ (18.44)} & \shortstack{' text'\\ (18.06)} & \shortstack{' int'\\ (17.22)} \\
\texttt{\bossymbol \newlinesymbol
def item(self, old, code, header, response, node, sub): \newlinesymbol
    """game phase birth \newlinesymbol
 \newlinesymbol
    :param header: cap session \newlinesymbol
    :param response: break player \newlinesymbol
    :param} & \texttt{\bossymbol \newlinesymbol
def image(self, key, file, filename, files, line, expected): \newlinesymbol
    """package crime framework \newlinesymbol
 \newlinesymbol
    :param host: dollar author \newlinesymbol
    :param command: cup spring \newlinesymbol
    :param} & 11.44 & \shortstack{' node'\\ (21.31)} & \shortstack{' code'\\ (17.17)} & \shortstack{' child'\\ (15.85)} & \shortstack{' file'\\ (20.06)} & \shortstack{' line'\\ (18.03)} & \shortstack{' node'\\ (17.91)} \\
\texttt{\bossymbol \newlinesymbol
def expected(self, root, results, host, module, names, files): \newlinesymbol
    """horse boot sector \newlinesymbol
 \newlinesymbol
    :param host: thinking rock \newlinesymbol
    :param module: rent tie \newlinesymbol
    :param} & \texttt{\bossymbol \newlinesymbol
def error(self, action, image, source, old, text, content): \newlinesymbol
    """charge conduct wife \newlinesymbol
 \newlinesymbol
    :param task: meaning shadow \newlinesymbol
    :param field: warning self \newlinesymbol
    :param} & 11.41 & \shortstack{' names'\\ (21.83)} & \shortstack{' name'\\ (19.86)} & \shortstack{' files'\\ (17.22)} & \shortstack{' image'\\ (17.93)} & \shortstack{' file'\\ (17.43)} & \shortstack{' name'\\ (17.00)} \\
\bottomrule
\end{tblr}

%% file: greaterthan-outputtable.tex
\begin{tblr}{X[l] X[l] c | c c c | c c c}
\toprule
      & & & \SetCell[c=3]{c} model &&& \SetCell[c=3]{c} circuit && \\
   input & patch input & loss & 1st   & 2nd & 3rd & 1st     & 2nd & 3rd \\
\midrule
The dispute lasted from the year 1694 to 16 & The voyage lasted from the year 1601 to 16 & 5.31 & \shortstack{'95'\\ (27.43)} & \shortstack{'97'\\ (26.22)} & \shortstack{'96'\\ (26.19)} & \shortstack{'95'\\ (27.09)} & \shortstack{'99'\\ (25.66)} & \shortstack{'97'\\ (25.32)} \\
The dispute lasted from the year 1694 to 16 & The expedition lasted from the year 1701 to 17 & 5.22 & \shortstack{'95'\\ (27.43)} & \shortstack{'97'\\ (26.22)} & \shortstack{'96'\\ (26.19)} & \shortstack{'95'\\ (27.07)} & \shortstack{'99'\\ (25.37)} & \shortstack{'97'\\ (24.50)} \\
The dispute lasted from the year 1694 to 16 & The pilgrimage lasted from the year 1601 to 16 & 5.18 & \shortstack{'95'\\ (27.43)} & \shortstack{'97'\\ (26.22)} & \shortstack{'96'\\ (26.19)} & \shortstack{'95'\\ (27.45)} & \shortstack{'99'\\ (26.14)} & \shortstack{'97'\\ (25.98)} \\
The dispute lasted from the year 1694 to 16 & The pilgrimage lasted from the year 1601 to 16 & 5.18 & \shortstack{'95'\\ (27.43)} & \shortstack{'97'\\ (26.22)} & \shortstack{'96'\\ (26.19)} & \shortstack{'95'\\ (27.45)} & \shortstack{'99'\\ (26.14)} & \shortstack{'97'\\ (25.98)} \\
The dispute lasted from the year 1694 to 16 & The voyage lasted from the year 1101 to 11 & 5.17 & \shortstack{'95'\\ (27.43)} & \shortstack{'97'\\ (26.22)} & \shortstack{'96'\\ (26.19)} & \shortstack{'95'\\ (24.12)} & \shortstack{'99'\\ (23.75)} & \shortstack{'50'\\ (23.06)} \\
The dispute lasted from the year 1694 to 16 & The voyage lasted from the year 1101 to 11 & 5.17 & \shortstack{'95'\\ (27.43)} & \shortstack{'97'\\ (26.22)} & \shortstack{'96'\\ (26.19)} & \shortstack{'95'\\ (24.12)} & \shortstack{'99'\\ (23.75)} & \shortstack{'50'\\ (23.06)} \\
The dispute lasted from the year 1694 to 16 & The voyage lasted from the year 1101 to 11 & 5.17 & \shortstack{'95'\\ (27.43)} & \shortstack{'97'\\ (26.22)} & \shortstack{'96'\\ (26.19)} & \shortstack{'95'\\ (24.12)} & \shortstack{'99'\\ (23.75)} & \shortstack{'50'\\ (23.06)} \\
The dispute lasted from the year 1694 to 16 & The pilgrimage lasted from the year 1201 to 12 & 5.15 & \shortstack{'95'\\ (27.43)} & \shortstack{'97'\\ (26.22)} & \shortstack{'96'\\ (26.19)} & \shortstack{'95'\\ (25.83)} & \shortstack{'99'\\ (25.40)} & \shortstack{'97'\\ (25.16)} \\
The raids lasted from the year 1788 to 17 & The expedition lasted from the year 1701 to 17 & 5.13 & \shortstack{'89'\\ (28.32)} & \shortstack{'90'\\ (27.60)} & \shortstack{'99'\\ (27.38)} & \shortstack{'89'\\ (28.30)} & \shortstack{'90'\\ (27.61)} & \shortstack{'93'\\ (27.17)} \\
The raids lasted from the year 1788 to 17 & The voyage lasted from the year 1601 to 16 & 5.12 & \shortstack{'89'\\ (28.32)} & \shortstack{'90'\\ (27.60)} & \shortstack{'99'\\ (27.38)} & \shortstack{'89'\\ (27.21)} & \shortstack{'90'\\ (27.16)} & \shortstack{'99'\\ (26.71)} \\
\bottomrule
\end{tblr}

%% file: percentiles-matched.tex
\begin{tblr}{lrrrrrr}
\toprule
   & \SetCell[c=2]{c} docstring  & & \SetCell[c=2]{c} greaterthan & & \SetCell[c=2]{c}{ioi} \\
 & \centering abs & \centering z-score & \centering abs & \centering z-score & \centering abs & \centering z-score \\
\midrule
count & 1000000.00 &  & 1000000.00 &  & 1000000.00 &  \\
mean & 1.38 &  & 0.08 &  & 0.59 &  \\
std & 0.93 &  & 0.06 &  & 0.53 &  \\
min & 0.01 & -1.46 & 0.01 & -1.29 & 0.00 & -1.10 \\
25\% & 0.71 & -0.72 & 0.04 & -0.66 & 0.21 & -0.71 \\
50\% & 1.11 & -0.28 & 0.07 & -0.25 & 0.42 & -0.32 \\
75\% & 1.78 & 0.43 & 0.10 & 0.32 & 0.80 & 0.39 \\
95\% & 3.29 & 2.04 & 0.20 & 1.94 & 1.65 & 1.98 \\
99\% & 4.57 & 3.41 & 0.33 & 3.99 & 2.49 & 3.57 \\
99.9\% & 6.06 & 5.01 & 0.46 & 6.14 & 3.85 & 6.12 \\
99.99\% & 7.33 & 6.37 & 0.54 & 7.45 & 5.34 & 8.92 \\
max & 9.97 & 9.19 & 0.60 & 8.47 & 8.83 & 15.47 \\
\bottomrule
\end{tblr}

%% file: ioi-outputtable-matched-corruptions.tex
\begin{tblr}{X[l] X[l] c | c c c | c c c}
\toprule
input & patch input & loss & model 0 & model 1 & model 2 & circuit 0 & circuit 1 & circuit 2 \\
\midrule
Then, Stephen and Jacob had a lot of fun at the house. Jacob gave a necklace to & Then, Jacob and Kelly had a lot of fun at the house. Adam gave a necklace to & 7.07 & \shortstack{' Stephen'\\ (17.77)} & \shortstack{' the'\\ (14.43)} & \shortstack{' his'\\ (14.32)} & \shortstack{' Jacob'\\ (20.96)} & \shortstack{' the'\\ (14.43)} & \shortstack{' his'\\ (14.27)} \\
Then, Alicia and Steven had a lot of fun at the hospital. Steven gave a kiss to & Then, Jacob and Jose had a lot of fun at the hospital. Amber gave a kiss to & 6.62 & \shortstack{' Alicia'\\ (18.22)} & \shortstack{' her'\\ (15.33)} & \shortstack{' the'\\ (15.10)} & \shortstack{' Steven'\\ (20.09)} & \shortstack{' her'\\ (14.66)} & \shortstack{' the'\\ (14.60)} \\
Then, Brandon and Rachel had a lot of fun at the store. Rachel gave a basketball to & Then, Rachel and Jesse had a lot of fun at the store. Paul gave a basketball to & 6.37 & \shortstack{' Brandon'\\ (18.98)} & \shortstack{' Rachel'\\ (15.06)} & \shortstack{' the'\\ (14.11)} & \shortstack{' Rachel'\\ (20.26)} & \shortstack{' the'\\ (14.15)} & \shortstack{' her'\\ (13.58)} \\
Then, Brandon and Rachel had a lot of fun at the garden. Rachel gave a kiss to & Then, Rebecca and Gregory had a lot of fun at the garden. Aaron gave a kiss to & 6.25 & \shortstack{' Brandon'\\ (18.37)} & \shortstack{' Rachel'\\ (16.05)} & \shortstack{' the'\\ (14.73)} & \shortstack{' Rachel'\\ (21.39)} & \shortstack{' the'\\ (15.49)} & \shortstack{' her'\\ (15.17)} \\
Then, Stephanie and Joseph had a lot of fun at the restaurant. Joseph gave a necklace to & Then, Joseph and Nathan had a lot of fun at the restaurant. Jennifer gave a necklace to & 6.19 & \shortstack{' Stephanie'\\ (19.24)} & \shortstack{' the'\\ (14.61)} & \shortstack{' her'\\ (13.82)} & \shortstack{' Joseph'\\ (19.28)} & \shortstack{' the'\\ (15.24)} & \shortstack{' them'\\ (14.32)} \\
Then, Patrick and Rachel had a lot of fun at the restaurant. Rachel gave a basketball to & Then, Samuel and Lauren had a lot of fun at the restaurant. Patrick gave a basketball to & 5.86 & \shortstack{' Patrick'\\ (18.38)} & \shortstack{' Rachel'\\ (14.38)} & \shortstack{' the'\\ (14.34)} & \shortstack{' Rachel'\\ (19.87)} & \shortstack{' the'\\ (14.48)} & \shortstack{' a'\\ (13.68)} \\
Then, Joshua and Rachel had a lot of fun at the garden. Rachel gave a kiss to & Then, Christina and Jonathan had a lot of fun at the garden. Melissa gave a kiss to & 5.75 & \shortstack{' Joshua'\\ (18.42)} & \shortstack{' Rachel'\\ (15.34)} & \shortstack{' the'\\ (15.12)} & \shortstack{' Rachel'\\ (20.42)} & \shortstack{' the'\\ (15.26)} & \shortstack{' her'\\ (14.71)} \\
Then, Vanessa and Stephen had a lot of fun at the garden. Stephen gave a kiss to & Then, Sara and Travis had a lot of fun at the garden. Rebecca gave a kiss to & 5.68 & \shortstack{' Vanessa'\\ (18.52)} & \shortstack{' the'\\ (15.26)} & \shortstack{' her'\\ (14.95)} & \shortstack{' Stephen'\\ (17.42)} & \shortstack{' the'\\ (15.46)} & \shortstack{' her'\\ (14.78)} \\
Then, Richard and Erin had a lot of fun at the store. Erin gave a ring to & Then, Allison and Jose had a lot of fun at the store. Nicholas gave a ring to & 5.63 & \shortstack{' Richard'\\ (17.33)} & \shortstack{' the'\\ (14.18)} & \shortstack{' Erin'\\ (13.54)} & \shortstack{' Erin'\\ (20.11)} & \shortstack{' the'\\ (14.36)} & \shortstack{' a'\\ (13.34)} \\
Then, Thomas and Dustin had a lot of fun at the store. Dustin gave a necklace to & Then, Allison and Jose had a lot of fun at the store. Amy gave a necklace to & 5.60 & \shortstack{' Thomas'\\ (16.60)} & \shortstack{' the'\\ (14.34)} & \shortstack{' his'\\ (13.40)} & \shortstack{' Dustin'\\ (19.55)} & \shortstack{' the'\\ (14.61)} & \shortstack{' Dust'\\ (13.71)} \\
\bottomrule
\end{tblr}

%% file: docstring-outputtable-matched-corruptions.tex
\begin{tblr}{X[l] X[l] c | c c c | c c c}
\toprule
input & patch input & loss & model 0 & model 1 & model 2 & circuit 0 & circuit 1 & circuit 2 \\
\midrule
\texttt{\bossymbol \newlinesymbol
def date(self, options, result, context, user, tag, error): \newlinesymbol
    """bench round model \newlinesymbol
 \newlinesymbol
    :param context: input sense \newlinesymbol
    :param user: album second \newlinesymbol
    :param} & \texttt{\bossymbol \newlinesymbol
def date(self, options, result, port, shape, new, error): \newlinesymbol
    """bench round model \newlinesymbol
 \newlinesymbol
    :param parent: input sense \newlinesymbol
    :param order: album second \newlinesymbol
    :param} & 8.73 & \shortstack{' tag'\\ (22.25)} & \shortstack{' tags'\\ (16.94)} & \shortstack{' context'\\ (16.52)} & \shortstack{' result'\\ (18.15)} & \shortstack{' error'\\ (16.81)} & \shortstack{':'\\ (16.53)} \\
\texttt{\bossymbol \newlinesymbol
def client(self, url, image, file, server, values, request): \newlinesymbol
    """fuel scale acid \newlinesymbol
 \newlinesymbol
    :param file: pub resident \newlinesymbol
    :param server: cell disk \newlinesymbol
    :param} & \texttt{\bossymbol \newlinesymbol
def client(self, url, image, token, code, state, request): \newlinesymbol
    """fuel scale acid \newlinesymbol
 \newlinesymbol
    :param content: pub resident \newlinesymbol
    :param msg: cell disk \newlinesymbol
    :param} & 7.64 & \shortstack{' values'\\ (23.30)} & \shortstack{' value'\\ (19.80)} & \shortstack{' data'\\ (17.54)} & \shortstack{' server'\\ (19.55)} & \shortstack{' file'\\ (18.60)} & \shortstack{' request'\\ (17.34)} \\
\texttt{\bossymbol \newlinesymbol
def source(self, content, group, project, tag, run, test): \newlinesymbol
    """seed post sample \newlinesymbol
 \newlinesymbol
    :param project: command distance \newlinesymbol
    :param tag: bank delay \newlinesymbol
    :param} & \texttt{\bossymbol \newlinesymbol
def source(self, content, group, results, options, name, test): \newlinesymbol
    """seed post sample \newlinesymbol
 \newlinesymbol
    :param default: command distance \newlinesymbol
    :param current: bank delay \newlinesymbol
    :param} & 7.54 & \shortstack{' run'\\ (21.77)} & \shortstack{' test'\\ (16.69)} & \shortstack{' project'\\ (16.68)} & \shortstack{' project'\\ (16.79)} & \shortstack{' target'\\ (16.23)} & \shortstack{' group'\\ (16.22)} \\
\texttt{\bossymbol \newlinesymbol
def check(self, action, last, text, base, run, table): \newlinesymbol
    """message duty scope \newlinesymbol
 \newlinesymbol
    :param text: bank height \newlinesymbol
    :param base: post sum \newlinesymbol
    :param} & \texttt{\bossymbol \newlinesymbol
def check(self, action, last, title, path, url, table): \newlinesymbol
    """message duty scope \newlinesymbol
 \newlinesymbol
    :param current: bank height \newlinesymbol
    :param call: post sum \newlinesymbol
    :param} & 7.43 & \shortstack{' run'\\ (20.48)} & \shortstack{' base'\\ (16.45)} & \shortstack{' line'\\ (16.41)} & \shortstack{' table'\\ (17.80)} & \shortstack{' str'\\ (17.56)} & \shortstack{' base'\\ (17.25)} \\
\texttt{\bossymbol \newlinesymbol
def call(self, path, end, option, log, instance, msg): \newlinesymbol
    """style drop demand \newlinesymbol
 \newlinesymbol
    :param option: colour entry \newlinesymbol
    :param log: impact cancer \newlinesymbol
    :param} & \texttt{\bossymbol \newlinesymbol
def call(self, path, end, task, update, new, msg): \newlinesymbol
    """style drop demand \newlinesymbol
 \newlinesymbol
    :param node: colour entry \newlinesymbol
    :param header: impact cancer \newlinesymbol
    :param} & 7.42 & \shortstack{' instance'\\ (20.95)} & \shortstack{' str'\\ (15.92)} & \shortstack{' bool'\\ (15.80)} & \shortstack{' log'\\ (17.91)} & \shortstack{' path'\\ (17.09)} & \shortstack{' str'\\ (16.92)} \\
\texttt{\bossymbol \newlinesymbol
def date(self, options, num, page, table, files, default): \newlinesymbol
    """root fund boy \newlinesymbol
 \newlinesymbol
    :param page: bar finger \newlinesymbol
    :param table: lane storm \newlinesymbol
    :param} & \texttt{\bossymbol \newlinesymbol
def date(self, options, num, value, config, order, default): \newlinesymbol
    """root fund boy \newlinesymbol
 \newlinesymbol
    :param valid: bar finger \newlinesymbol
    :param group: lane storm \newlinesymbol
    :param} & 7.29 & \shortstack{' files'\\ (23.10)} & \shortstack{' file'\\ (21.28)} & \shortstack{' filename'\\ (17.47)} & \shortstack{' table'\\ (17.95)} & \shortstack{' num'\\ (16.62)} & \shortstack{' str'\\ (16.54)} \\
\texttt{\bossymbol \newlinesymbol
def tag(self, content, port, test, end, model, count): \newlinesymbol
    """top release drop \newlinesymbol
 \newlinesymbol
    :param test: collection reading \newlinesymbol
    :param end: protein dream \newlinesymbol
    :param} & \texttt{\bossymbol \newlinesymbol
def tag(self, content, port, date, target, text, count): \newlinesymbol
    """top release drop \newlinesymbol
 \newlinesymbol
    :param string: collection reading \newlinesymbol
    :param index: protein dream \newlinesymbol
    :param} & 7.26 & \shortstack{' model'\\ (19.99)} & \shortstack{' test'\\ (16.13)} & \shortstack{' models'\\ (15.54)} & \shortstack{' string'\\ (16.73)} & \shortstack{' int'\\ (16.02)} & \shortstack{' bool'\\ (15.57)} \\
\texttt{\bossymbol \newlinesymbol
def instance(self, state, size, project, image, fields, run): \newlinesymbol
    """father sort horse \newlinesymbol
 \newlinesymbol
    :param project: dollar protein \newlinesymbol
    :param image: duty net \newlinesymbol
    :param} & \texttt{\bossymbol \newlinesymbol
def instance(self, state, size, server, end, target, run): \newlinesymbol
    """father sort horse \newlinesymbol
 \newlinesymbol
    :param config: dollar protein \newlinesymbol
    :param description: duty net \newlinesymbol
    :param} & 7.12 & \shortstack{' fields'\\ (21.71)} & \shortstack{' field'\\ (18.85)} & \shortstack{' name'\\ (16.97)} & \shortstack{' value'\\ (14.27)} & \shortstack{' str'\\ (14.22)} & \shortstack{' int'\\ (14.15)} \\
\texttt{\bossymbol \newlinesymbol
def data(self, parent, new, url, model, found, count): \newlinesymbol
    """bone trip user \newlinesymbol
 \newlinesymbol
    :param url: user location \newlinesymbol
    :param model: device object \newlinesymbol
    :param} & \texttt{\bossymbol \newlinesymbol
def data(self, parent, new, date, order, message, count): \newlinesymbol
    """bone trip user \newlinesymbol
 \newlinesymbol
    :param field: user location \newlinesymbol
    :param command: device object \newlinesymbol
    :param} & 7.12 & \shortstack{' found'\\ (19.72)} & \shortstack{' discovered'\\ (15.77)} & \shortstack{' data'\\ (15.38)} & \shortstack{' url'\\ (16.38)} & \shortstack{' description'\\ (15.97)} & \shortstack{' model'\\ (15.96)} \\
\texttt{\bossymbol \newlinesymbol
def user(self, current, server, table, tag, result, group): \newlinesymbol
    """cake saving pub \newlinesymbol
 \newlinesymbol
    :param table: fashion user \newlinesymbol
    :param tag: committee tree \newlinesymbol
    :param} & \texttt{\bossymbol \newlinesymbol
def user(self, current, server, fields, base, match, group): \newlinesymbol
    """cake saving pub \newlinesymbol
 \newlinesymbol
    :param order: fashion user \newlinesymbol
    :param old: committee tree \newlinesymbol
    :param} & 7.07 & \shortstack{' result'\\ (22.19)} & \shortstack{' user'\\ (16.85)} & \shortstack{' current'\\ (16.37)} & \shortstack{' table'\\ (16.64)} & \shortstack{' user'\\ (16.43)} & \shortstack{' server'\\ (16.14)} \\
\bottomrule
\end{tblr}

%% file: greaterthan-outputtable-matched-corruptions.tex
\begin{tblr}{X[l] X[l] c | c c c | c c c}
\toprule
input & patch input & loss & model 0 & model 1 & model 2 & circuit 0 & circuit 1 & circuit 2 \\
\midrule
The sanctions lasted from the year 1520 to 15 & The sanctions lasted from the year 1501 to 15 & 0.60 & \shortstack{'30'\\ (25.73)} & \shortstack{'25'\\ (25.17)} & \shortstack{'40'\\ (24.95)} & \shortstack{'21'\\ (25.54)} & \shortstack{'23'\\ (25.27)} & \shortstack{'22'\\ (25.15)} \\
The sanctions lasted from the year 1520 to 15 & The sanctions lasted from the year 1501 to 15 & 0.60 & \shortstack{'30'\\ (25.73)} & \shortstack{'25'\\ (25.17)} & \shortstack{'40'\\ (24.95)} & \shortstack{'21'\\ (25.54)} & \shortstack{'23'\\ (25.27)} & \shortstack{'22'\\ (25.15)} \\
The reforms lasted from the year 1520 to 15 & The reforms lasted from the year 1501 to 15 & 0.59 & \shortstack{'30'\\ (25.91)} & \shortstack{'25'\\ (25.26)} & \shortstack{'40'\\ (25.14)} & \shortstack{'21'\\ (25.38)} & \shortstack{'23'\\ (25.07)} & \shortstack{'22'\\ (24.98)} \\
The accord lasted from the year 1520 to 15 & The accord lasted from the year 1501 to 15 & 0.58 & \shortstack{'30'\\ (25.75)} & \shortstack{'25'\\ (25.24)} & \shortstack{'40'\\ (24.78)} & \shortstack{'21'\\ (25.85)} & \shortstack{'22'\\ (25.39)} & \shortstack{'23'\\ (25.30)} \\
The accord lasted from the year 1520 to 15 & The accord lasted from the year 1501 to 15 & 0.58 & \shortstack{'30'\\ (25.75)} & \shortstack{'25'\\ (25.24)} & \shortstack{'40'\\ (24.78)} & \shortstack{'21'\\ (25.85)} & \shortstack{'22'\\ (25.39)} & \shortstack{'23'\\ (25.30)} \\
The accord lasted from the year 1520 to 15 & The accord lasted from the year 1501 to 15 & 0.58 & \shortstack{'30'\\ (25.75)} & \shortstack{'25'\\ (25.24)} & \shortstack{'40'\\ (24.78)} & \shortstack{'21'\\ (25.85)} & \shortstack{'22'\\ (25.39)} & \shortstack{'23'\\ (25.30)} \\
The flights lasted from the year 1580 to 15 & The flights lasted from the year 1501 to 15 & 0.56 & \shortstack{'90'\\ (27.01)} & \shortstack{'85'\\ (24.96)} & \shortstack{'80'\\ (24.83)} & \shortstack{'90'\\ (26.09)} & \shortstack{'83'\\ (25.51)} & \shortstack{'85'\\ (25.41)} \\
The flights lasted from the year 1580 to 15 & The flights lasted from the year 1501 to 15 & 0.56 & \shortstack{'90'\\ (27.01)} & \shortstack{'85'\\ (24.96)} & \shortstack{'80'\\ (24.83)} & \shortstack{'90'\\ (26.09)} & \shortstack{'83'\\ (25.51)} & \shortstack{'85'\\ (25.41)} \\
The flights lasted from the year 1680 to 16 & The flights lasted from the year 1601 to 16 & 0.54 & \shortstack{'90'\\ (28.71)} & \shortstack{'80'\\ (26.84)} & \shortstack{'85'\\ (26.71)} & \shortstack{'90'\\ (27.09)} & \shortstack{'83'\\ (26.79)} & \shortstack{'82'\\ (26.49)} \\
The flights lasted from the year 1680 to 16 & The flights lasted from the year 1601 to 16 & 0.54 & \shortstack{'90'\\ (28.71)} & \shortstack{'80'\\ (26.84)} & \shortstack{'85'\\ (26.71)} & \shortstack{'90'\\ (27.09)} & \shortstack{'83'\\ (26.79)} & \shortstack{'82'\\ (26.49)} \\
\bottomrule
\end{tblr}